\documentclass[10pt,twocolumn,letterpaper]{article}

\usepackage{format_iccv2023/iccv}
\usepackage{times}
\usepackage{epsfig}
\usepackage{graphicx}
\usepackage{amsmath}
\usepackage{amssymb}

\usepackage{subfigure}
\usepackage{multirow}
\usepackage{subfigure}
\usepackage{pifont}
\usepackage{amsfonts}
\usepackage{bm}

\newcommand{\cmark}{\ding{51}}

\newcommand{\R}{\mathbb{R}}

\usepackage[pagebackref=true,breaklinks=true,letterpaper=true,colorlinks,bookmarks=false]{hyperref}

\usepackage[capitalize]{cleveref}
\crefname{section}{Sec.}{Secs.}
\Crefname{section}{Section}{Sections}
\crefname{table}{Tab.}{Tabs.}
\Crefname{table}{Table}{Tables}

\iccvfinalcopy 

\def\iccvPaperID{5054} 

\begin{document}

\title{SemARFlow: Injecting Semantics into Unsupervised Optical Flow Estimation for Autonomous Driving}

\author{Shuai Yuan, Shuzhi Yu, Hannah Kim, and Carlo Tomasi\\
Duke University\\
{\tt\small \{shuai, shuzhiyu, hannah, tomasi\}@cs.duke.edu}
}

\maketitle
\ificcvfinal\thispagestyle{empty}\fi

\begin{abstract}
Unsupervised optical flow estimation is especially hard near occlusions and motion boundaries and in low-texture regions. We show that additional information such as semantics and domain knowledge can help better constrain this problem. We introduce SemARFlow, an unsupervised optical flow network designed for autonomous driving data that takes estimated semantic segmentation masks as additional inputs. This additional information is injected into the encoder and into a learned upsampler that refines the flow output. In addition, a simple yet effective semantic augmentation module provides self-supervision when learning flow and its boundaries for vehicles, poles, and sky. Together, these injections of semantic information improve the KITTI-2015 optical flow test error rate from 11.80\% to 8.38\%. We also show visible improvements around object boundaries as well as a greater ability to generalize across datasets. Code is available at \url{https://github.com/duke-vision/semantic-unsup-flow-release}.
\end{abstract}



\section{Introduction} \label{sec:intro}


Optical flow estimation, \ie, pixel-level motion tracking across video frames, has broad applications in many computer vision tasks that include object tracking~\cite{shin2005optical}, video editing~\cite{gao2020flow, kim2022cross}, and autonomous driving~\cite{capito2020optical, shen2023optical}.

Thanks to the success of deep convolutional neural networks~\cite{krizhevsky2017imagenet,he2016deep,li2021survey} and transformer networks~\cite{dosovitskiy2020image,touvron2021training,zhu2020deformable,khan2022transformers} in computer vision, many top-performing supervised optical flow networks have been proposed in recent years~\cite{dosovitskiy2015flownet,sun2018pwc,hur2019iterative,teed2020raft,zhang2021separable}, in which ground-truth labels supervise training. However, real optical flow is hard to label, so most supervised methods train (or at least pre-train) on synthetic datasets~\cite{dosovitskiy2015flownet,mayer2016large,butler2012naturalistic,Richter_2017} which makes them hard to adapt to real applications due to the significant gap between synthetic and real data~\cite{tremblay2018training,kim2022spatially}.

Due to label scarcity, \emph{unsupervised} training of optical flow estimators~\cite{yu2016back} instead uses loss terms based on the assumptions of constant brightness and smooth flow~\cite{meister2018unflow,jonschkowski2020matters,luo2021upflow}. Some self-supervision techniques have also been studied to enhance model performance~\cite{liu2019ddflow,liu2019selflow,liu2020learning}. Unsupervised training makes it possible to train flow networks directly on large real datasets from the target domain.

\begin{figure}[tb]
\begin{center}
    \includegraphics[width=\linewidth]{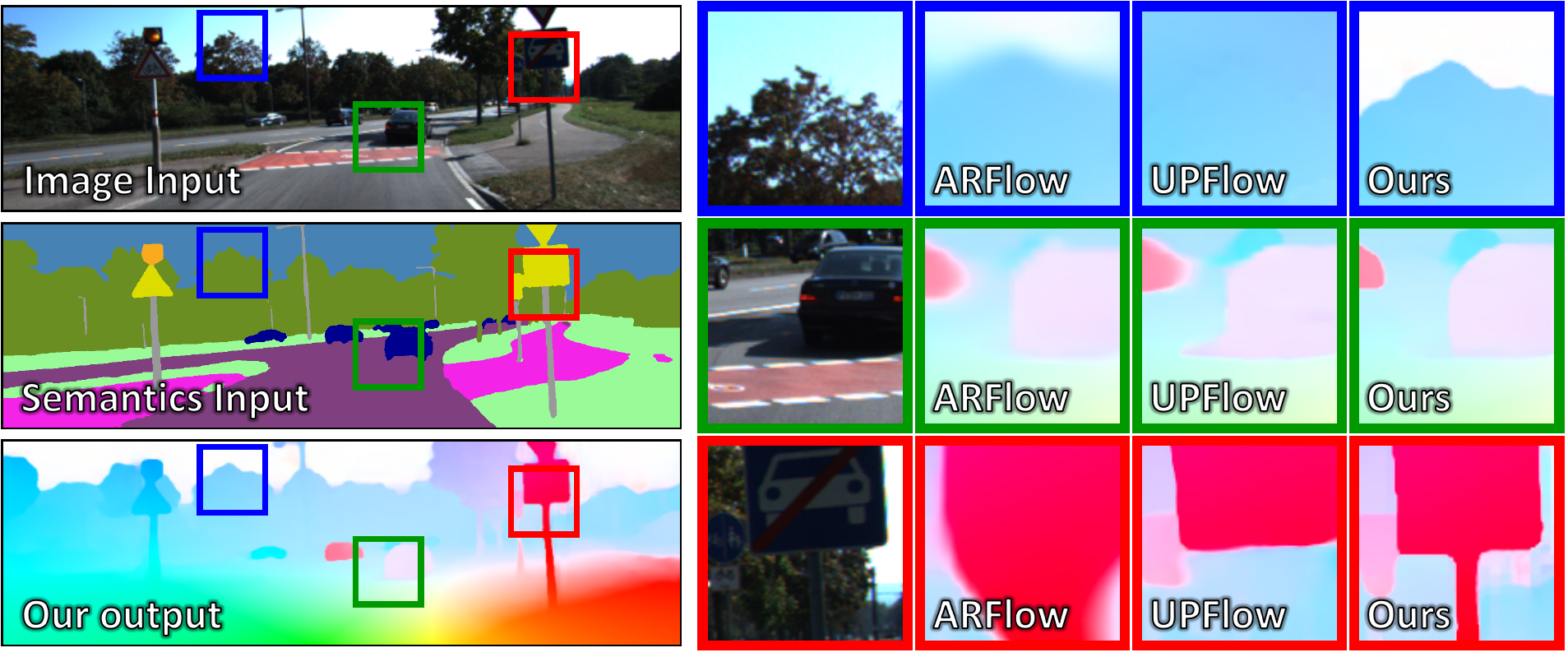}
    \caption{An example on KITTI-2015 test set (sample \#63). Only one input frame is shown for conciseness. Our SemARFlow takes additional semantic segmentation inputs (estimated by an off-the-shelf model) and outputs much sharper flow around semantic boundaries.}
    \label{fig:intro_demo}
\end{center}
\end{figure}

Even so, unsupervised optical flow estimation is a poorly constrained problem. Brightness is not constant in regions with occlusions~\cite{wang2018occlusion} or on shiny surfaces~\cite{marsal2023brightflow}, nor is it smooth across motion boundaries~\cite{kim2021joint,yu2022unsupervised}. Moreover, points in regions with poor texture content~\cite{jianbo1994good} or in dark shadows~\cite{shen2023optical} are difficult to track as they are easily confused with neighboring points (the so-called aperture problem). This dearth of reliable constraints makes the unsupervised training of flow networks especially challenging.

A natural way to address this issue is to inject additional constraints in the form of semantics and domain knowledge. For example, in autonomous driving applications~\cite{yurtsever2020survey,capito2020optical}, we have clear expectations about object types and layouts of the scene, as well as prior knowledge of how each type of object typically moves. We focus on this application domain and show that this additional information helps the network achieve better flow results.

To make an autonomous driving system work well in reality, it is best to train it on real data. However, annotating real driving video with optical flow labels is expensive~\cite{yuan2022optical}, as it requires careful synchronization and calibration of diverse sensors including cameras, LiDAR, and GPS/IMU~\cite{geiger2012we}, aided by some manual annotation and curation based on CAD models of moving objects~\cite{kitti15}.

In contrast, annotating semantic labels seems much more feasible, and indeed semantic labels are available in most (if not all) existing driving datasets. We consider semantic segmentation because it provides semantics at the pixel level, the same level as optical flow. As one of the most popular and well-studied tasks in modern computer vision, semantic segmentation~\cite{long2015fully,noh2015learning,zheng2021rethinking} has been extensively adopted for autonomous driving systems. In this paper, we show that adding semantic segmentation inputs helps improve unsupervised optical flow performance significantly.

Specifically, we first infer semantic segmentation maps using an off-the-shelf model~\cite{zhu2019improving}, which of course is trained with semantic labels. An encoder with semantic map input is used to aggregate image and semantic features (\cref{subsec:enc}), and a learned upsampler is added into the iterative decoder to refine flow around object boundaries given semantic inputs (\cref{subsec:dec}). We also propose a simple yet effective semantic augmentation module for self-supervision, which provides realistic augmentations specific to the vehicles, poles, and sky classes based on domain knowledge and segmentation maps (\cref{subsec:aug}). An occluder cache is implemented to improve efficiency (\cref{subsec:cache}). Semantic augmentation provides challenging samples for self-supervision, which help train flow better in occluded regions and around foreground object boundaries. 

Overall, by injecting semantic segmentation inputs, our SemARFlow network achieves significantly better unsupervised flow results both quantitatively (\cref{subsec:test}) and qualitatively (\cref{subsec:test_demo}). Adapted from ARFlow~\cite{liu2020learning}, SemARFlow reduces KITTI-2015~\cite{kitti15} test error from 11.80\% to 8.38\% and out-performs the current unsupervised state-of-the-art UPFlow~\cite{luo2021upflow} (9.38\%) by a clear margin. The example in \cref{fig:intro_demo} also demonstrates visible improvements around object boundaries. The effectiveness of each module is justified through an extensive ablation study (\cref{subsec:ablation}) and improvement analysis (\cref{subsec:impr}). In addition to performance boost, unsupervised flow networks with additional semantic inputs generalize better across different datasets (\cref{subsec:gen}).

Our research is essentially novel compared to previous approaches. Some early work incorporates semantics in traditional energy-minimization methods~\cite{lucas1981iterative,horn1981determining} for optical flow through geometric constraints such as piece-wise rigid motion and planar surface motion ~\cite{sevilla2016optical,hur2016joint,bai2016exploiting,wulff2017optical}. In comparison, to the best of our knowledge, we are the first to inject semantics into the unsupervised training of recent optical flow networks. Some research also trains a network for segmentation and optical flow jointly~\cite{ding2020every}. In contrast, we leverage existing, separately trained segmentation systems both because they are available and because modularity---separating segmentation from flow estimation---is important for the development of large real-application systems such as autonomous driving.

In summary, our contributions are as follows. 
\begin{itemize}
    \item To the best of our knowledge, we are the first to explore adding semantic inputs to assist the unsupervised training of deep optical flow networks. 
    \item We propose a simple yet effective network called SemARFlow that achieves state-of-the-art results both quantitatively and qualitatively. Our model works well on real-life occlusions and yields sharp motion boundaries around objects.
    \item We provide full training and inference code as well as trained models to encourage follow-up research.
\end{itemize}


\section{Related work} \label{sec:rel_work}

\paragraph{Supervised optical flow} 


Since the introduction of the pioneering deep optical flow network FlowNet~\cite{dosovitskiy2015flownet}, more and more top-performing CNN-based flow estimators have been proposed over the years~\cite{Ilg2017FlowNet2E,Ranjan2017OpticalFE,sun2018pwc,hui2018liteflownet,hur2019iterative,teed2020raft}. Recently, vision transformer networks and attention mechanism have also been applied to this problem, and these have achieved state-of-the-art performance on benchmark datasets~\cite{xu2021high,jiang2021learning,zhang2021separable,sui2022craft,xu2022gmflow}. The supervised methods are often pre-trained on large synthetic datasets such as FlyingChairs~\cite{dosovitskiy2015flownet} and FlyingThings3D~\cite{mayer2016large} before fine-tuning on the target dataset. However, there is a clear gap between the artificially generated data and real scenarios.

\paragraph{Unsupervised optical flow} 

Due to the lack of ground-truth labels, unsupervised optical flow estimation uses surrogate losses such as photometric loss and smoothness loss to supervise training~\cite{yu2016back,ren2017unsupervised}. To tackle the issues around occlusion regions, various methods have been proposed including estimated occlusion masks~\cite{wang2018occlusion}, forward-backward consistency~\cite{meister2018unflow}, and multi-frame fusion~\cite{janai2018unsupervised,ren2019fusion}. To better upsample the flow in the decoder, UPFlow~\cite{luo2021upflow} additionally predicts a confidence map and an interpolation flow to guide flow refinement and has become the state-of-the-art unsupervised flow method.

Some latest research has also explored the use of self-supervision to enhance flow prediction. Early methods apply the knowledge distillation technique to train a two-stage teacher-student network~\cite{liu2019ddflow,liu2019selflow}. ARFlow~\cite{liu2020learning} further improves this idea by generating reliable self-supervision signals from data transformations, while merging the two training stages into single-stage training with one added loss term. SimFlow~\cite{im2020unsupervised} replaces the handcrafted features with deep self-supervised features to measure similarity in the unsupervised losses. SMURF~\cite{stone2021smurf} utilizes a RAFT-like structure~\cite{teed2020raft} and applies multi-frame self-supervised training with many technical improvements. Our SemARFlow also uses self-supervision but with the guidance of semantic segmentation, which is much more realistic than guidance without semantics. 

\paragraph{Semantic segmentation}  

Semantic segmentation classifies each pixel of the given image into semantic objects. Fully Convolutional Network (FCN)~\cite{long2015fully} is one of the early CNN-based segmentation methods. It takes inputs of arbitrary sizes and outputs dense pixel-level predictions, becoming one of the popular backbone architectures for follow-up work. Deconvolution network~\cite{noh2015learning} is also proposed to better recover low-level details of the prediction. One main challenge for these systems is that they lack global scene information. Subsequent work addresses this issue by enlarging the receptive field of the network with global pyramid pooling layers as in PSPNet~\cite{zhao2017pyramid}, hybrid dilated convolutions~\cite{wang2018understanding}, and a fast-down-sampling strategy~\cite{yu2018bisenet,yu2021bisenet}.  Attention modules have also shown to help in semantic segmentation by capturing full-image dependencies of all pixels~\cite{huang2019ccnet,zheng2021rethinking} or the semantic inter-dependencies across spatial and channel dimensions~\cite{fu2019dual}.

There also exists extensive work on semantic segmentation in the context of autonomous driving~\cite{borse2021inverseform,tao2020hierarchical,ganeshan2021warp,cai2021multi,zhu2019improving}, thanks to the publication of large-scale driving datasets~\cite{cordts2016cityscapes,kitti15}. To better train the network on a coarsely labeled dataset like KITTI~\cite{kitti15}, Zhu~\etal use a video prediction network SDCNet~\cite{reda2018sdc} to synthesize new training samples with relaxed label propagation~\cite{zhu2019improving}. Due to their good performance on KITTI~\cite{kitti15}, we utilize their network models to infer semantic inputs for all our experiments.

\paragraph{Combining semantics and optical flow}  

Though there has been much progress on both semantic segmentation and optical flow estimation, the semantic optical flow problem (how to exploit semantics to help optical flow estimation) has received limited attention in recent years, and the current best results are thus much outdated. Some early methods incorporate semantics through geometric constraints to refine flow on various semantic regions, such as planar regions (using homographies)~\cite{sevilla2016optical}, static regions (using rigid camera motion and epipolar constraints)~\cite{hur2016joint,wulff2017optical}, and rigid objects (estimating rigid motion for each object instance)~\cite{bai2016exploiting}. However, most methods are traditional flow methods based on energy minimization, where an initial flow estimate is usually needed and semantics is mostly used for refinement. In comparison, we explore adapting latest unsupervised optical flow networks to leverage semantic inputs in one single stage of estimation.

Apart from using semantics to help flow, some research has also explored using optical flow to help semantic segmentation~\cite{rashed2019optical,li2019flow2seg}, or to train both tasks jointly~\cite{ding2020every}. There are also studies on exploiting semantics on some other correspondence matching tasks such as stereo matching~\cite{wu2019semantic,yang2018segstereo} and 3D scene flow estimation, where some additional depth cues such as stereo camera inputs~\cite{ren2017cascaded,feng2021ss} or point clouds~\cite{shi2022safit} are needed.

\section{Method} \label{sec:method}

\begin{figure*}[tb]
\begin{center}
\subfigure[shared pyramidal encoder ($i\in\{1, 2\}$)]{
    \includegraphics[width=0.33\linewidth]{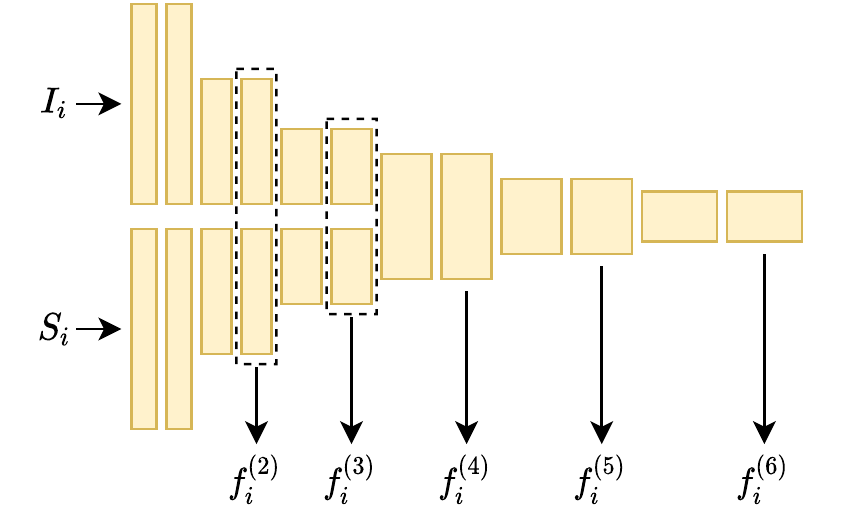}
    \label{fig:enc}
    }
\subfigure[one iteration (at the $l$th-level) of the iterative decoder ($l\in\{6, 5, 4, 3, 2\}$)]{
    \includegraphics[width=0.63\linewidth]{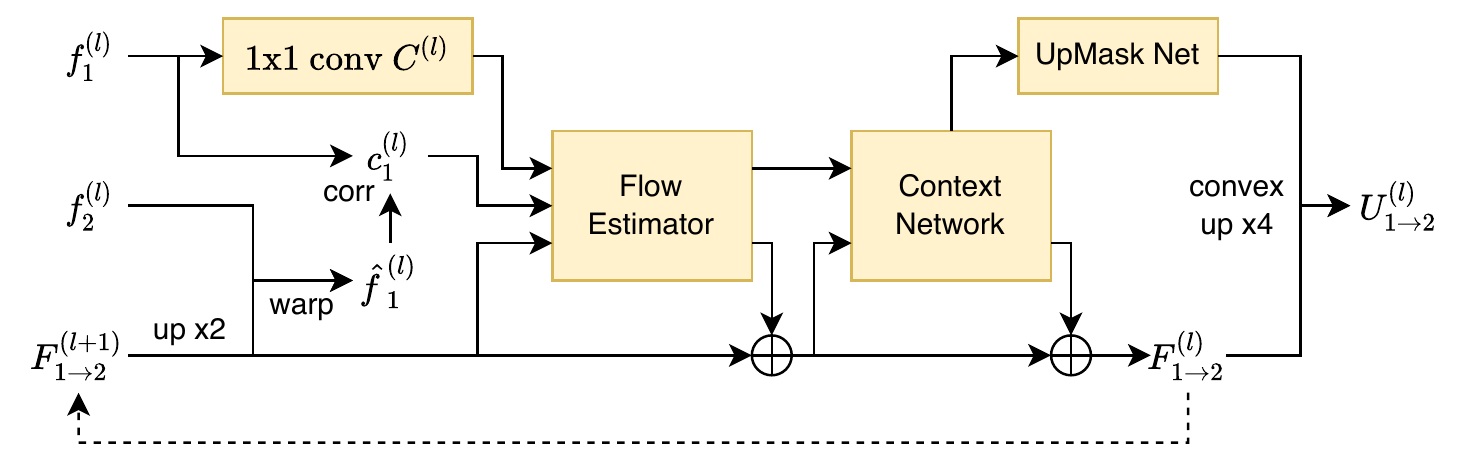}
    \label{fig:dec}
    }
\end{center}
\caption{Network structure. See text in \cref{subsec:enc} and \cref{subsec:dec} for explanations. More detailed diagrams are in appendix.}
\label{fig:enc_dec}
\end{figure*}

\begin{figure*}[tb]
\begin{center}
    \includegraphics[width=\linewidth]{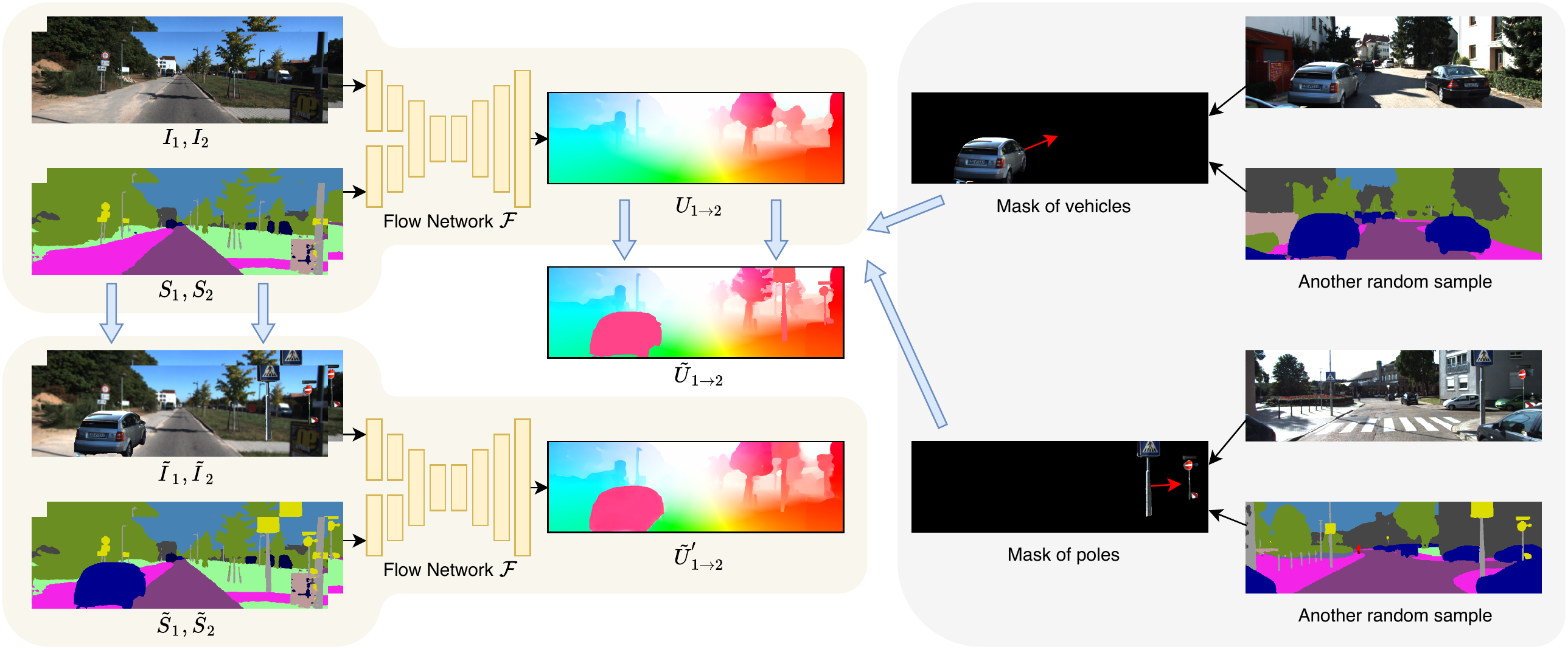}
    \caption{Illustration of semantic augmentation as self-supervision. See text in \cref{subsec:aug} for details.}
    \label{fig:sem_aug_demo}
\end{center}
\end{figure*}

Our network is adapted from the two-frame version of ARFlow~\cite{liu2020learning}, which uses a light-weight PWCNet~\cite{sun2018pwc} as its backbone. The inputs are two consecutive frames $I_1, I_2\in\R^{H\times W\times 3}$ as well as their semantic segmentation maps $S_1, S_2\in\{0, 1, \cdots, c\}^{H\times W}$, where the number of classes $c$ is 19 as we use the Cityscapes format~\cite{cordts2016cityscapes}.  A detailed diagram of the structure of our network can be found in the Appendix.

\subsection{Semantic encoder} \label{subsec:enc}

We first inject semantic information into the encoder. As shown in \cref{fig:enc}, shared by each frame $i\in\{1, 2\}$, separate convolutional layers extract features from image $I_i$ and semantics $S_i$ (one-hot encoded). Features from the two pipelines are concatenated and fed to additional layers. Features at different resolutions $(H/2^l, W/2^l)$ form a pyramid $\{f_i^{(l)}\mid 2\leq l\leq 6\}$. The semantic information in these features helps delineate objects in dark shadows, where appearance is more homogeneous.

\subsection{Iterative decoder with a learned upsampler} \label{subsec:dec}

Following ~\cite{hur2019iterative,liu2020learning}, an iterative residual refinement decoder starts from zero estimate $F_{1\to 2}^{(7)}=0$. For iteration $l\in\{6, 5, 4, 3, 2\}$), the decoder refines feature map $F_{1\to 2}^{(l+1)}$ into $F_{1\to 2}^{(l)}$ based on features $f_1^{(l)}, f_2^{(l)}$ (\cref{fig:dec}).

More specifically, $F_{1\to 2}^{(l+1)}$ is upsampled to match the resolution and used to warp $f_2^{(l)}$ to yield warped feature $\hat f_1^{(l)}$. Correlation volumes are computed between $f_1^{(l)}$ and $\hat f_1^{(l)}$. A one-by-one convolutional layer $C^{(l)}$ compresses the number of channels to a fixed number so that the same layer can be reused across all iterations as proposed by ~\cite{hur2019iterative}. A flow estimator network predicts a flow residual to be added to the current estimate, and a context network then aggregates flow information spatially and refines the current flow again.

A \emph{learned} upsampler network upsamples the final output $F_{1\to2}^{(2)}$ in our system. This is different from ARFlow~\cite{liu2020learning}, which simply uses four-fold bilinear interpolation, making the final flow boundaries blurry. In contrast, our model learns to sharpen flow boundaries based on the semantic inputs, which have clear boundaries around moving objects. To this end, we add a convex upsampler network similar to the one in RAFT~\cite{teed2020raft}. Different from UPFlow~\cite{luo2021upflow}, our learned upsampler is only used to upsample from internal estimate $F_{1\to 2}^{(l)}$ to output flow $U_{1\to 2}^{(l)}$, but not when we upsample $F_{1\to 2}^{(l+1)}$ at the first step in the decoder iteration. The upsampled $U_{1\to 2}^{(l)}$ has resolution $(H/2^{l-2}, W/2^{l-2})$, so $U_{1\to 2}=U_{1\to 2}^{(2)}$ (with the original resolution) is the final flow prediction of the network.

\subsection{Semantic augmentation as self-supervision}  \label{subsec:aug}

ARFlow has a very effective in-network augmentation module that samples random transformations $\mathcal{T}_{\theta_1}, \mathcal{T}_{\theta_2}$ of the flow prediction $U_{1 \to2}$ in the first pass of the network and then uses the transformed images $\hat I_1=\mathcal{T}_{\theta_1}(I_1), \hat I_2 = \mathcal{T}_{\theta_2}(I_2)$ in a second pass. The prediction $U_{1\to2}$ is also transformed accordingly and used to self-supervise the output of the second pass. See ~\cite{liu2020learning} for details.

We retain the augmentation module but make a third pass of the network using semantics-transformed inputs for self-supervision in addition to the ARFlow augmentations of appearance (\eg, color jitter, random noise) and spatial transformations (\eg, random rotation, random rescaling). The idea behind semantic augmentation is to blend in real object motions across samples.

To this end (see \cref{fig:sem_aug_demo}), we carve objects from other samples based on their semantic maps and paste them as moving foreground objects into the current sample $I_1, I_2$, thereby producing augmented images $\tilde{I}_1, \tilde I_2$. The segmentation maps $S_1, S_2$ and the first pass output flow $U_{1\to2}$ are transformed accordingly, and the transformed $\tilde U_{1\to2}$ self-supervises the third-pass network outputs $\tilde U'_{1\to2}=\mathcal{F}(\tilde I_1, \tilde I_2, \tilde S_1, \tilde S_2)$, where $\mathcal{F}$ is the flow network. 

Similar to occlusion hallucination in SelFlow~\cite{liu2019selflow}, our semantic augmentation also creates new occlusions and uses the reliable non-occluded flow from the first pass to self-supervise the flow on those newly occluded pixels. However, rather than superpixels filled with noise, we create occlusions with real objects, which are much more realistic.

Moreover, since we determine the motion of each occluder, we can use its known optical flow for motion self-supervision. As illustrated in \cref{fig:sem_aug_demo}, the occluder motion can be very different from that of the background. This trains the model to assume less smoothness around  highly-dynamic objects and to output sharper motion estimates near motion boundaries. These hallucinated but realistic occluders also provide more examples of actively moving objects, which are relatively scarce in the original datasets.

\paragraph{Vehicles} We augment vehicle classes (car, truck, bus, train) as they are major sources of errors in previous models (see \cref{tab:impr} for statistics). We group these four classes because they are easily confused with each other in the semantic input. To carve out a single car instance with high probability, we find connected components of vehicle segmentation regions using OpenCV~\cite{opencv_library} and accept masks whose bounding boxes are 50-300 pixels wide and 50-150 pixels high. We also drop masks that occupy less than 60\% area of their bounding boxes to remove largely occluded vehicles.

\paragraph{Poles} We augment pole classes (pole, traffic light, traffic sign) not only because they are common foreground objects that cause occlusions, but also because their motions are usually poorly estimated due to their thin shapes. We use a 200-pixel wide, image-height sliding window to scan the segmentation of these pole classes and find the window with the most pole pixels. If  more than 10\% pole pixels occur in it, that object mask is stored. This produces groups of nearby poles rather than just individual poles, which may not provide enough augmentation by themselves.

\paragraph{Sky} We also augment the sky through prior knowledge that flow in the sky should be very small. We shrink the sky flow estimate by half before using it for self-supervision. As shown in \cref{fig:sem_aug_demo}, the middle tree part in $U_{1\to2}$ is very blurry, with a large part of sky mixed in there. After shrinking sky flow by half, the tree in $\tilde U_{1\to2}$ now has sharper flow around the boundary between tree and sky. Using this as self-supervision helps the model sharpen the motion boundaries between sky and other regions.

Another way to apply this prior knowledge is to post-process, \ie, shrink the final \emph{output} flow in the sky region, instead of using the shrunk flow to self-supervise. However, the latter approach is preferable because it allows the network to balance prior knowledge (self-supervision) with the current sample observation (photometric loss). Our experiments do show cases where some small objects (such as some overhead electric power lines) are misclassified as part of the sky in the semantic input. In these cases, the network learns to rely less on prior knowledge. 

\subsection{Occluder cache} \label{subsec:cache}

During training, we maintain a cache of occluding objects for semantic augmentation. We sample objects from the cache to augment the current batch and then push any new objects found in the batch into the cache, with random replacement if the cache is full.

Specifically, after we load a sample for training, we first use the current model to infer its optical flow and then search within the sample for any cars (or poles) that meet the standards defined in \cref{subsec:aug} as ``cutouts''. For each cutout, we first find the flow vectors on all pixels that belong to that cutout and then compute the mean flow for later reference. We store the current sample, the cutout mask, and the mean flow together in the cache.

Later on, when we randomly select a cutout to augment another sample, we also retrieve the stored mean flow of this cutout, which is then augmented by random rescaling (by 0.8-1.5 times) and reversing (with 50\% probability). We use the augmented mean flow to translate the entire cutout object as an occluder to generate a new sample. For realism, occluders are pasted to the same location they occupied in their image of origin. Holding occluders in a cache provides a random mixture of samples across multiple batches and makes semantic augmentation very efficient.

\begin{table*}[tb]
\begin{center}
\begin{tabular}{cl|cc|cccccc}
\hline
\multicolumn{2}{c|}{\multirow{3}{*}{Method}} & \multicolumn{2}{c|}{Train}              & \multicolumn{6}{c}{Test}                                      \\ \cline{3-10}
\multicolumn{2}{c|}{}                       & \multicolumn{1}{c|}{2012} & \multicolumn{1}{c|}{2015} & \multicolumn{2}{c|}{2012}         & \multicolumn{4}{c}{2015}        \\
\multicolumn{2}{c|}{}                       & \multicolumn{1}{c|}{EPE}  & EPE             & \underline{Fl-noc} & \multicolumn{1}{c|}{EPE} & \underline{Fl-all} & Fl-noc & Fl-bg & Fl-fg \\ \hline
\multicolumn{1}{c|}{\multirow{4}{*}{\rotatebox[origin=c]{90}{supervised}}}     & PWC-Net+ ~\cite{sun2019models}         & \multicolumn{1}{c|}{-}     & (1.50)  &  3.36    & \multicolumn{1}{c|}{1.4}    & 7.72  &  4.91    &   7.69   & 7.88      \\
\multicolumn{1}{c|}{}                         & IRR-PWC~\cite{hur2019iterative}           & \multicolumn{1}{c|}{-}    & (1.63)     & 3.21 & \multicolumn{1}{c|}{1.6} & 7.65  & 4.86 & 7.68 & 7.52  \\
\multicolumn{1}{c|}{}                           & RAFT~\cite{teed2020raft}              & \multicolumn{1}{c|}{-}    & (0.63)     & -    & \multicolumn{1}{c|}{-}   & 5.10  & 3.07 & 4.74 & 6.87  \\
\multicolumn{1}{c|}{}                           & Separable Flow~\cite{zhang2021separable}     & \multicolumn{1}{c|}{-}    & (0.69) & -    & \multicolumn{1}{c|}{-}   & 4.53  & 2.78 & 4.25 & 5.92  \\ \hline
\multicolumn{1}{c|}{\multirow{8}{*}{\rotatebox[origin=c]{90}{unsupervised}}} & SelFlow~\cite{liu2019selflow}            & \multicolumn{1}{c|}{1.69} & 4.84                     & 4.31   & \multicolumn{1}{c|}{2.2} & 14.19  & 9.65   & 12.68 & 21.74 \\
\multicolumn{1}{c|}{}                          & SimFlow~\cite{im2020unsupervised}            & \multicolumn{1}{c|}{-} & 5.19                     & -   & \multicolumn{1}{c|}{-} & 13.38  & 8.21   & 12.60 & 17.27 \\
\multicolumn{1}{c|}{}                           & ARFlow ~\cite{liu2020learning}           & \multicolumn{1}{c|}{1.44} & 2.85       & -    & \multicolumn{1}{c|}{1.8} & 11.80 & -    & -    & -     \\
\multicolumn{1}{c|}{}                           & UFlow ~\cite{jonschkowski2020matters}            & \multicolumn{1}{c|}{1.68} & 2.71       & 4.26 & \multicolumn{1}{c|}{1.9} & 11.13 & 8.41 & 9.78 & 17.87 \\
\multicolumn{1}{c|}{}                           & UPFlow ~\cite{luo2021upflow}           & \multicolumn{1}{c|}{\textbf{1.27}} & 2.45      & -    & \multicolumn{1}{c|}{\textbf{1.4}} & 9.38  & -    & -    & -     \\ \cline{2-10} 
\multicolumn{1}{c|}{}                           & Ours (baseline) & \multicolumn{1}{c|}{ 1.39} &   2.61                   &  4.30   & \multicolumn{1}{c|}{1.7 } & 9.89  & 6.98   & 8.82 & 15.21 \\
\multicolumn{1}{c|}{}                         & Ours (+enc)$^\dagger$ & \multicolumn{1}{c|}{1.29} & 2.42                    & 3.97   & \multicolumn{1}{c|}{1.5} & 8.99 & 3.97   & 8.19  & 13.01 \\
\multicolumn{1}{c|}{}                         & Ours (+enc +aug)$^\dagger$  & \multicolumn{1}{c|}{1.28} & \textbf{2.18}                   & \textbf{3.90}   & \multicolumn{1}{c|}{1.5} & \textbf{8.38} & \textbf{3.90}    & \textbf{7.48}  & \textbf{12.91} \\ \hline
\end{tabular}
\end{center}
\caption{KITTI benchmark results (EPE/px and Fl/\%). Metrics evaluated at `all' (all pixels, default for EPE), `noc' (non-occlusions), `bg' (background), and `fg' (foreground). Key metrics (used in official ranking) are underlined. Our `baseline' is an adapted ARFlow with added learned upsampler and no smoothness loss. `+enc` means adding semantic encoder; `+aug' means adding semantic augmentation. `()' means evaluation data used in training. `-' means unavailable. $^\dagger$ denotes models with semantic inputs. For all metrics, lower is better.}
\label{tab:unsup_test}
\end{table*}

\subsection{Loss functions}

\paragraph{Photometric loss} In the first pass, forward and backward flow $U^{(l)}_{1\to2}, U^{(l)}_{2\to1}$ at each level are predicted and occlusion masks $O^{(l)}_{1\to2}, O^{(l)}_{2\to1}$ are computed by forward-backward consistency~\cite{meister2018unflow}. Frames are warped by
\[
I_i'^{(l)}(\bm p) = I_j^{(l)}(\bm p + U_{i\to j}^{(l)}(\bm p)),\quad (i, j\in\{1, 2\})
\]
where $I_i^{(l)}$ is $I_i$ down-sampled to the $l$-th scale, and $\bm p$ denotes pixel coordinates at that scale. Following~\cite{liu2020learning},  three measures, namely L$_1$-distance ($\rho_1$), structural similarity (SSIM)~\cite{wang2004image} ($\rho_2$), and census loss~\cite{meister2018unflow} ($\rho_3$) are linearly combined to measure photometric differences between $I_i^{(l)}$ and $I_i'^{(l)}$. Occlusion regions are masked out and both forward and backward directions are taken into account at each level. The final photometric loss is
\begin{equation} \label{eq:ph_loss_one_scale}
    \ell_{\text{ph}} = \frac{1}{2}\sum_{(i, j)\in \{(1, 2), (2, 1)\}}\sum_{l=2}^6 a_l\sum_{k=1}^3 c_k \rho_k(I_i^{(l)}, I_i'^{(l)}, O_{i\to j}^{(l)}).
\end{equation} 

\paragraph{Smoothness loss} Unlike most previous work, we do not include a smoothness loss because we find it to conflict with our learned upsampler. See ablation study in \cref{subsec:ablation}.

\paragraph{Augmentation loss} As in ARFlow~\cite{liu2020learning}, our second forward pass computes the L$_1$-distance between the transformed flow $\hat U_{1\to 2}$ and the second pass output $\hat U'_{1\to 2}=\mathcal{F}(\hat I_1, \hat I_2, \hat S_1, \hat S_2)$ for each pixel $\bm p$ as
\[
\hat D(\bm p) = \|\hat{U}_{1\to2}(\bm p) - \hat{U}'_{1\to2}(\bm p)\|_1.
\]
This loss is then averaged over the transformed non-occluded region where self-supervision $\hat U_{1\to 2}$ is accurate
\begin{equation}
    \ell_{\text{ar}} = \frac{\sum_{\bm p} (1- \hat O_{1\to 2}(\bm p)) \hat D(\bm p)}{\sum_{\bm p} (1- \hat O_{1\to 2}(\bm p))}.
\end{equation}

For semantic augmentation (third pass), since we are also doing self-supervision with a new generated sample, we use a similar loss definition
\begin{equation}
    \ell_{\text{aug}} = \frac{\sum_{\bm p} (1- \tilde O_{1\to 2}(\bm p)) \tilde D(\bm p)}{\sum_{\bm p} (1- \tilde O_{1\to 2}(\bm p))},
\end{equation}
where $\tilde D(\bm p)$ is the distance between the semantic-augmented flow $\tilde U_{1\to2}$ and the third-pass output $\tilde U'_{1\to2}$
\[
\tilde D(\bm p) = \|\tilde{U}_{1\to2}(\bm p) - \tilde{U}'_{1\to2}(\bm p)\|_1,
\]
and the augmented mask $\tilde O_{1\to2}$ is now computed by
\[
\tilde O_{1\to2} = 1 - \max(1 - O_{1\to2}, M)
\]
because we penalize on both the originally non-occluded region $1 - O_{1\to2}$ and the pasted foreground object mask $M$ (of which we know the true motion).

\paragraph{Final loss} (with $\lambda=0.02$ as in ARFlow~\cite{liu2020learning}):
\begin{equation}
\ell = \ell_{\text{ph}} + \lambda (\ell_{\text{ar}} + \ell_{\text{aug}})\;.
\end{equation}

\section{Experiments}\label{sec:exp}

\subsection{Datasets}
We mainly use KITTI~\cite{kitti12,kitti15} and Cityscapes~\cite{cordts2016cityscapes} datasests for experiments. Following~\cite{liu2020learning}, we train our model first on KITTI raw sequences~\cite{kitti15} (55.7k samples) and then fine-tune on KITTI-2015 multi-view extension~\cite{kitti15} (11.8k samples). We validate our model using KITTI-2015 train~\cite{kitti15} (200 samples) and KITTI-2012 train set~\cite{kitti12} (194 samples) since they are the only sets that have optical flow labels. We also train on Cityscapes~\cite{cordts2016cityscapes} sequences (83.3k samples) to test model generalization ability, and we sample every other frame to match its frame rate with KITTI. All semantic segmentation inputs are estimated using an off-the-shelf model~\cite{zhu2019improving} with a DeepLabV3Plus~\cite{chen2018encoder} backbone, which achieves 83.45\% mean IoU on Cityscapes and 72.82\% mean IoU on KITTI.

\subsection{Implementation details}
We implement the model in PyTorch~\cite{NEURIPS2019_9015} \footnote{Code and instructions are available at \url{https://github.com/duke-vision/semantic-unsup-flow-release}.}. We use the Adam optimizer~\cite{kingma2014adam} with $\beta_1=0.9, \beta_2=0.999$ and batch size 4. We first train on KITTI raw sequences~\cite{kitti15} for 100k iterations with a fixed learning rate 0.0002 and then train on KITTI multi-view extension set~\cite{kitti12,kitti15} for another 100k iterations using OneCycleLR schedule~\cite{smith2019super} with maximum learning rate 0.0004 and linear annealing.

For data augmentation, we include random horizontal flipping and swapping of the input frames. We resize the inputs to $256\times 832$ before feeding into the network. The photometric loss weight for each scale $a_l$ $(2\leq l\leq 6)$ in \cref{eq:ph_loss_one_scale} are set as 1, 1, 1, 1, 0. The weights for three photometric distance measures $\rho_k$ $(1\leq k\leq 3)$ in \cref{eq:ph_loss_one_scale} are set as 0.15, 0.85, 0 for the first 50k iterations and 0, 0, 1 afterwards. We start the appearance and spatial augmentation (second pass as in ARFlow~\cite{liu2020learning}) at 50k iterations, and semantic augmentation (third pass) after 150k iterations. 

\subsection{Benchmark testing results} \label{subsec:test}

As common practice, we evaluate optical flow predictions based on two error measurements, Fl (error rate) and EPE (mean L$_\text{2}$ distance). When computing Fl, the estimate of each pixel is considered correct if the error is smaller than 3 pixels or 5\% of the magnitude of ground-truth flow~\cite{kitti15}.

As shown in \cref{tab:unsup_test}, our semantic modules (both semantic encoder and semantic augmentation) improve performance on both KITTI-2012 and KITTI-2015 sets on all metrics. Our final model achieves 8.38\% Fl-all error rate on KITTI-2015 test set, which is significantly better than ARFlow~\cite{liu2020learning} (11.80\%), from which we adapt. We also outperform the current state-of-the-art UPFlow~\cite{luo2021upflow} (9.38\%) by a clear margin. All these results strongly suggest that adding semantic inputs can help improve unsupervised flow estimation significantly.

\begin{table}[tb]
\begin{center}
\begin{tabular}{l|cccc}
\hline
\multicolumn{1}{c|}{Method} & \underline{Fl-all} & Fl-noc & Fl-bg & Fl-fg \\ \hline
JFS~\cite{hur2016joint}                         & 17.07  & 9.81   & 15.9  & 22.92 \\
SOF~\cite{sevilla2016optical}                         & 16.81  & 10.86  & 14.63 & 27.73 \\
MRFlow~\cite{wulff2017optical}                  & 12.19 & 8.86  & 10.13     & 22.52 \\
Bai~\etal~\cite{bai2016exploiting}                            & 11.62  & 8.75   & 8.61  & 26.69 \\
Ours (final)                      & \textbf{8.38}   & \textbf{3.90}    & \textbf{7.48}  & \textbf{12.91} \\ \hline
\end{tabular}
\end{center}
\caption{KITTI-2015 test results (Fl/\%) compared with other semantic optical flow methods. Metrics evaluated at `all' (all pixels), `noc' (non-occlusions), `bg' (background), and `fg' (foreground).}
\label{tab:sem_flow_test}
\end{table}

In addition, \cref{tab:sem_flow_test} shows that our test results significantly outperform all previous semantic optical flow methods, most of which are based on traditional energy minimization. We are the first to apply semantic inputs to recent unsupervised flow networks, so we are able to push the state-of-the-art by a clear margin.

\subsection{Qualitative results} \label{subsec:test_demo}



\begin{figure*}[tb]
\begin{center}
    \includegraphics[width=\linewidth]{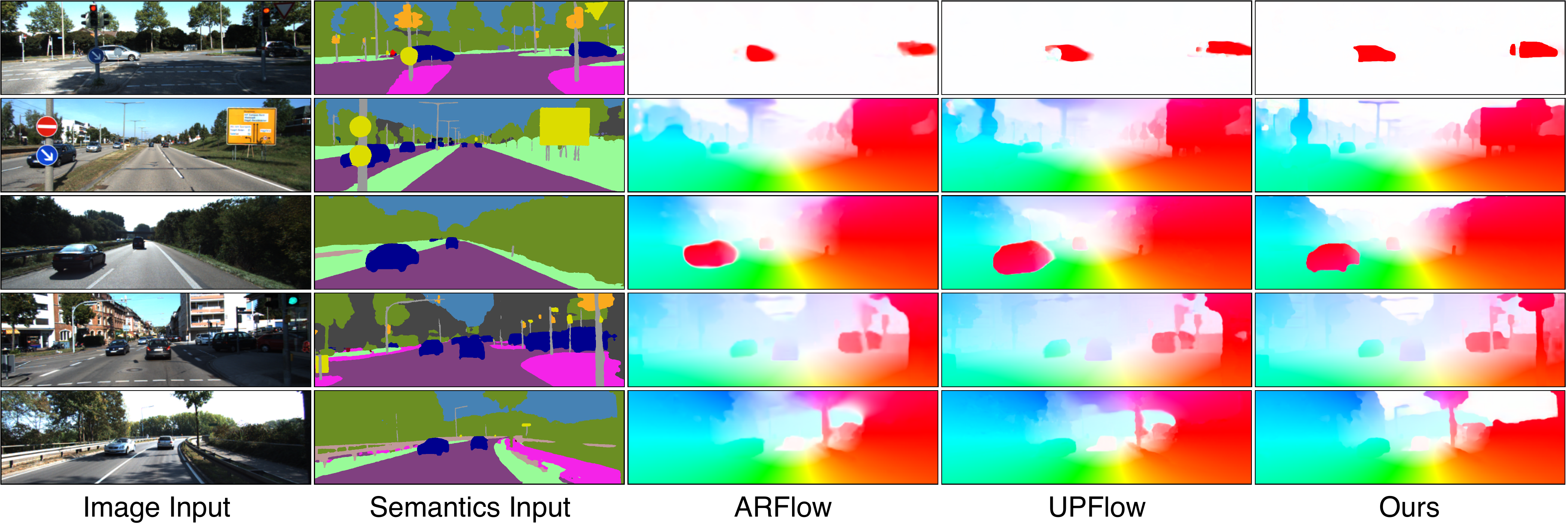}
    \caption{Qualitative results on KITTI test set (sample \#7, 20, 38, 112, 183) compared with ARFlow~\cite{liu2020learning} and UPFlow~\cite{luo2021upflow}. We only show first frame input for conciseness. Our outputs have much clearer boundaries around objects. For more qualitative examples, see Appendix.}
    \label{fig:expr_demo}
\end{center}    
\end{figure*}

Some qualitative results are shown in \cref{fig:expr_demo}. We can see that our flow outputs have very sharp boundaries and very clean object motion, especially around cars and poles. Our model successfully learns to adapt flow estimation to the semantic map input. Moreover, our model is able to handle very challenging samples where the foreground car motion is drastically different from the background due to the challenging samples we create in semantic augmentation.

\subsection{Ablation study}\label{subsec:ablation}

\setlength{\tabcolsep}{4pt}
\begin{table}[tb]
\begin{center}
\small
\begin{tabular}{cccc|cccc}
\hline
up  & no sm & enc & aug & \underline{Fl-all} & EPE-all & EPE-noc & EPE-occ \\ \hline
    &       &         &         & 10.36  & 2.90   & 2.07   & 6.97   \\ \hline
    &       & 1       &         & 9.92   & 2.69   & 1.89   & 6.57   \\
    &       & 2       &         & 9.83   & 2.65   & 1.86   & 6.40   \\
    &       & 3       &         & 9.75  & 2.61   & 1.85   & 6.54   \\
    &       & 4       &         & 9.75  & 2.64   & 1.85   & 6.55   \\ \hline
\cmark &       &         &         & 10.22  & 2.83   & 1.97    & 6.75   \\
\cmark & \cmark   &         &         & 8.87  & 2.61   & 1.85   & 6.04   \\ \hline
\cmark & \cmark   & 3       &         & 8.26   & 2.42   & 1.73   & 5.79   \\
\cmark & \cmark   &         & \cmark     &  8.80      &  2.48       &  1.60       &  6.64       \\
\cmark & \cmark   & 3       & \cmark     & \textbf{7.79}  & \textbf{2.18}   & \textbf{1.40}   & \textbf{5.64}   \\ \hline
\end{tabular}
\end{center}
\caption{Ablation study on KITTI-2015 train set (EPE/px and Fl/\%). `up' means the learned upsampler; `no sm' means turning off smoothness loss; `enc' means the level at which we merge semantic features and image features in the encoder; `aug' means semantic augmentation. }
\label{tab:ablation}
\end{table}
\setlength{\tabcolsep}{6pt}

We also do ablation study to show the effectiveness of each of our proposed modules. We can see in \cref{tab:ablation} that adding a semantic encoder to the vanilla ARFlow does help, and the optimal number of encoder layer groups added is 3. Also, our learned upsampler improves results significantly once we turn off the smoothness loss. This makes sense because the upsampler learns to adapt to boundaries, which can be distracted by the smoothness loss. Moreover, both our semantic encoder and semantic augmentation module help improve the results further, which suggests the effectiveness of adding semantic inputs.

\setlength{\tabcolsep}{5pt}
\begin{table}[tb]
\begin{center}
\begin{tabular}{l|cccc}
\hline
\multicolumn{1}{c|}{Options of aug}  & \underline{Fl-all} & EPE-all & EPE-noc & EPE-occ \\ \hline
Ours (final)                      & \textbf{7.79}   & \textbf{2.18}   & \textbf{1.40}   & 5.64   \\ \hline
start from 100k               & 7.92  & 2.19   & \textbf{1.40}   & \textbf{5.41}   \\ 
vehicles only                 & 7.94   & 2.21   & 1.44   & 5.78   \\
loss on new occ               & 8.15  & 2.27   & 1.42   & 5.62    \\ \hline
\end{tabular}
\end{center}
\caption{Ablation study of different semantic augmentation options on KITTI-2015 train (EPE/px and Fl/\%). See text for explanations.}
\label{tab:ablation_aug}
\end{table}
\setlength{\tabcolsep}{6pt}

Another ablation study is on different options for the semantic augmentation (\cref{tab:ablation_aug}). We try starting augmentation earlier from 100k iterations, which works slightly worse. There are mainly two reasons: (1) we use the first forward output to self-supervise our augmented output, so if we start early, the model will use poor output to self-supervise the augmented pass; and (2) our semantic augmentation creates very challenging samples with objects moving very differently from the background, so exposing these hard samples to the model too early may make it hard to train. Apart from this, we also try only using vehicles to augment as well as focusing loss on the newly occluded region, which are both inferior to our final version.

\subsection{Improvement analysis} \label{subsec:impr}

\begin{table*}[tb]
\begin{center}
\begin{tabular}{c|cccccccc}
\hline
       & road    & car     & terrain & vegetation & sidewalk & building & wall    & pole    \\ \hline
Proportion     & 42.4\%   & 17.6\%   & 14.0\%   & 11.6\%      & 6.2\%     & 4.1\%     & 1.3\%    & 1.1\%    \\ \hline
ARFlow~\cite{liu2020learning} & 4.57    & 15.79   & 9.42    & 15.34      & 4.61     & 6.00     & 16.34   & 10.75   \\
Ours (final)   & 3.67    & 10.17   & 8.74    & 13.47      & 3.09     & 4.27     & 11.18   & 9.44    \\ \hline
Relative improvement      & 19.6\% & 35.6\% & 7.3\%  & 12.2\%    & {33.0\%}  & {28.7\%}  & {31.6\%} & 12.2\% \\
Reweighed contribution       & \textcolor{red}{19.0\%} & \textcolor{red}{49.4\%} & 4.8\%  & \textcolor{red}{10.9\%}    & {4.7\%}  & {3.5\%}  & {3.4\%} & 0.7\% \\ \hline
\end{tabular}
\end{center}
\caption{KITTI-2015 flow error (Fl-all/\%) for each semantic class. The first row shows the proportion of each class among evaluated pixels (KITTI flow does not evaluate on every pixel). We only show classes that account for at least 1\% of the evaluated pixels.}
\label{tab:impr}
\end{table*}

It may be interesting to understand where our improvements come from in terms of semantic classes, so we compute the KITTI-2015 train set error for each class. As shown in \cref{tab:impr}, our model improves the flow on every class. After reweighing the absolute improvement of each class based on their proportion in the evaluated pixels, we find that car flow accounts for nearly half of our improvement on Fl-all, indicating the effectiveness of augmenting vehicle objects. 

\subsection{Generalization ability} \label{subsec:gen}
We also investigate the generalization ability of our semantic-aided flow models. We train flow models on Cityscapes and directly test them on KITTI-2015 train set without fine-tuning. Following~\cite{ranjan2019competitive}, we crop the bottom 25\% of the frame to remove the car logo and resize the frame to size $256\times704$. All other experiment settings are exactly the same as for KITTI. As we can see from \cref{tab:city}, adding semantic inputs significantly helps our unsupervised flow model generalize and adapt better across datasets.

\setlength{\tabcolsep}{3pt}
\begin{table}[tb]
\begin{center}
\begin{tabular}{l|cccc}
\hline
\multicolumn{1}{c|}{Method} & \underline{Fl-all} & EPE-all & EPE-noc & EPE-occ \\ \hline
ARFlow (our impl.)                     & 13.21 & 4.08  & 2.88  & 9.40 \\
Ours (baseline)                            & 12.27  & 3.81   & 2.43   & 9.91   \\
Ours (+enc)$^\dagger$                              & 11.28  & 3.33   & 2.12   & 8.75    \\
Ours (+enc +aug)$^\dagger$                         & \textbf{10.32}  & \textbf{2.64}     & \textbf{1.56}     & \textbf{7.12}   \\ \hline
\end{tabular}
\end{center}
\caption{Generalization results (train on Cityscapes, and test on KITTI-2015 train). We use our implementation of ARFlow~\cite{liu2020learning} to run this experiment.  Our `baseline' is an adapted ARFlow with added learned upsampler and no smoothness loss. `+enc` means adding semantic encoder; `+aug' means adding semantic augmentation. $^\dagger$ denotes models with semantic inputs. }
\label{tab:city}
\end{table}
\setlength{\tabcolsep}{6pt}


\section{Conclusion and future work} \label{sec:con}

In this paper, we show that adding semantic segmentation inputs can help significantly improve the performance of unsupervised optical flow networks on autonomous driving datasets. We propose a novel network model called SemARFlow, a semantic adaptation of ARFlow~\cite{liu2020learning}, with a semantic encoder, a learned upsampler, and a semantic augmentation module, where some domain-specific motion prior has been taken into account. Our network better predicts flow at occlusion regions and effectively sharpens flow estimates around object boundaries, even for very challenging samples. The additional semantic inputs also make our network generalize better across datasets.

\begin{figure}[tb]
\begin{center}
    \includegraphics[width=\linewidth]{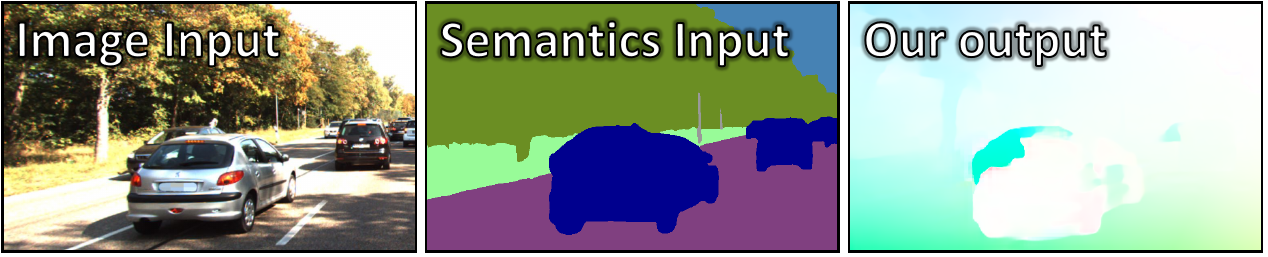}
    \caption{An example on KITTI-2015 train set (sample \#164). See the discussion in \cref{sec:con} for details.}
    \label{fig:con_demo}
\end{center}
\end{figure}

One direction for future work is motivated by the example illustrated in \cref{fig:con_demo}. Our model does output sharp flow boundaries between the cars and the background, thanks to good boundary information from the semantic maps. However, the boundary between these two fast-moving cars is inferred from image information only, because their semantic maps merge into one. As a result, that part of the flow boundary is less accurate. An interesting direction for future work is to use instance-level semantics as input for more detailed object masks.

\paragraph{Acknowledgments.} This research is based upon work supported by the National Science Foundation under Grant No. 1909821.


{\small
\bibliographystyle{format_iccv2023/ieee_fullname}
\bibliography{main}
}

\clearpage
\appendix
\section{Network structure}

\begin{figure*}[tb]
\begin{center}
    \includegraphics[width=0.7\linewidth]{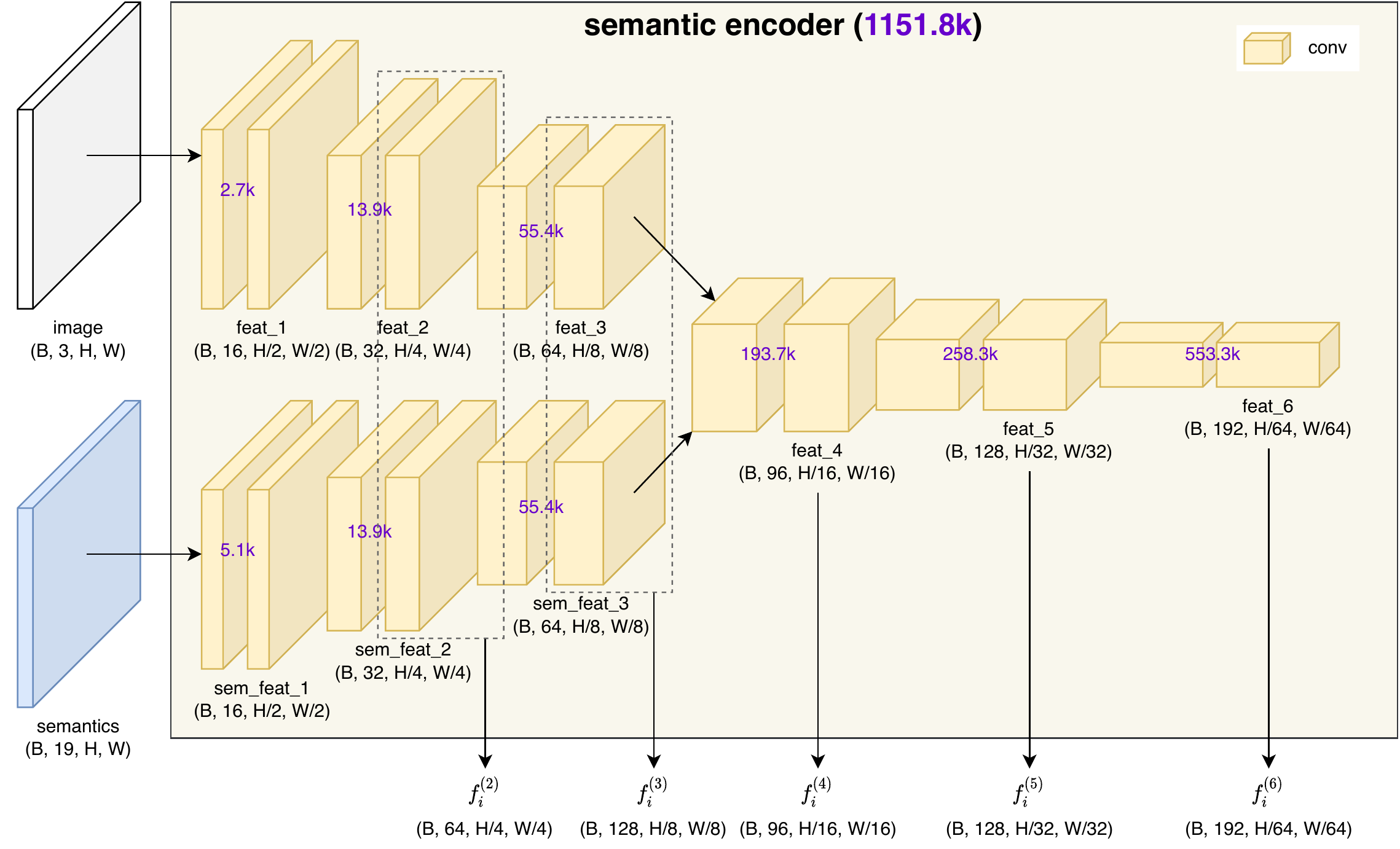}
    \caption{Semantic encoder network structure. Purple numbers show the number of parameters. $B$ is batch size; $(H, W)$ is the input resolution.}
    \label{fig:enc_app}
\end{center} 
\end{figure*}

\begin{figure*}[tb]
\begin{center}
    \includegraphics[width=\linewidth]{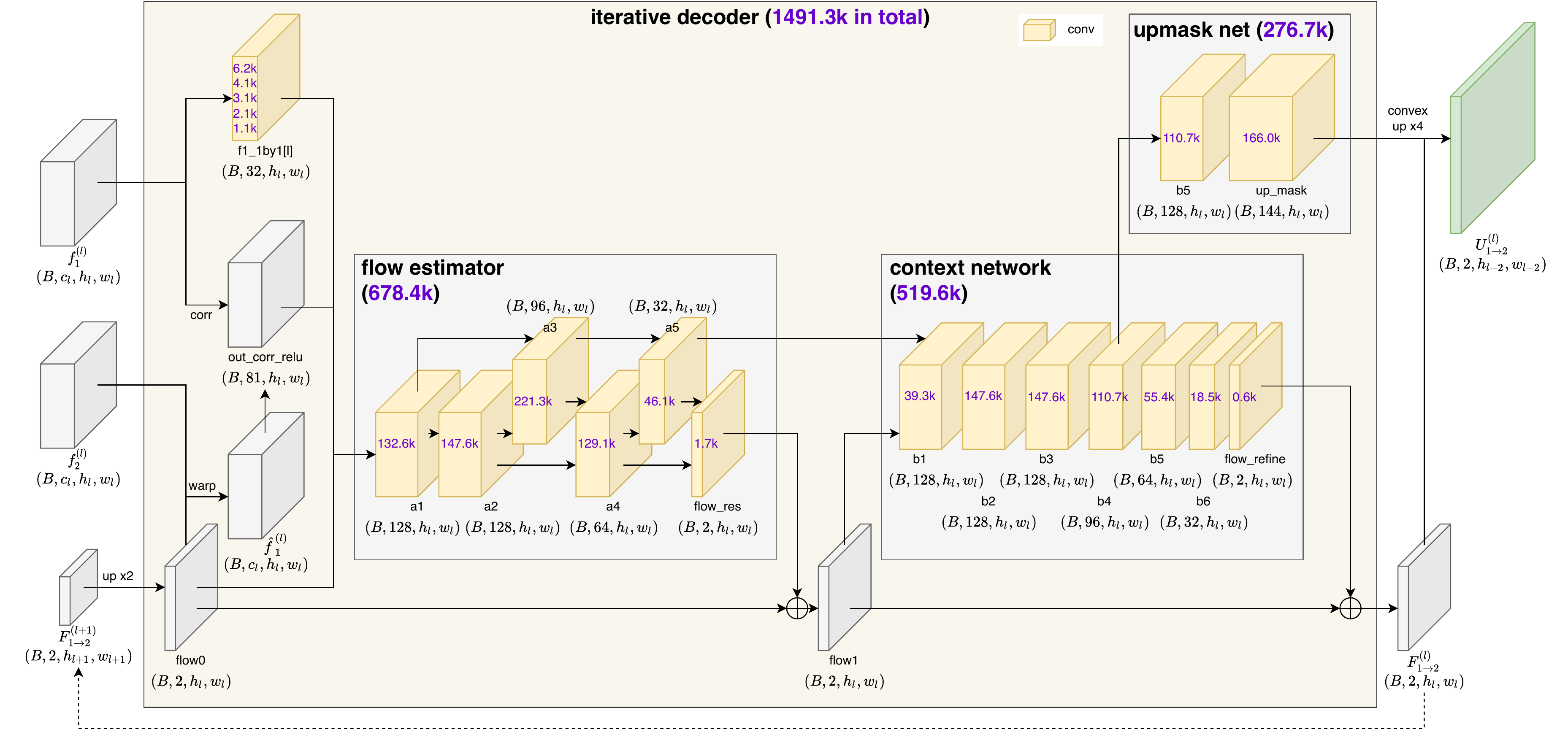}
    \caption{Iterative decoder network structure. Purple numbers show the number of parameters. The $1\times1$ convolution layers are not shared, so we show the number of parameters at each level. $B$ is batch size; $(h_l, w_l)$ is the resolution of the $l$-th level; $c_l$ is the number of channels. }
    \label{fig:dec_app}
\end{center}
\end{figure*}

Our detailed network structure is shown in \cref{fig:enc_app} and \cref{fig:dec_app}. The dimension of each convolutional layer and each tensor in the network is marked in the figure. 

In the semantic encoder, we have six levels of layer groups, and each layer group consists of two convolutional layers with leaky ReLU non-linearity. The layers in the same group have the same number of output channels marked in the figure. We use $3\times 3$ convolutional kernels everywhere. The dilation value is set as 2 whenever we need to reduce the dimension by half.

The iterative decoder has a flow estimator, a context network, and a upmask net, which are shared across all levels. Some $1\times 1$ convolutional layers are applied to transform the input features of different levels to features with the same number of channels so that they can be fed into the same shared flow estimator.

Our upmask net outputs a 144-channel mask for upsampling. We first unfold the 144 channels to 16 groups, each of which has 9 values. Since we are upsampling four times, each original value needs to correspond to 16 values in the output, and each output value is computed as a convex linear combinition of the $3\times 3$ input window, so each group of 9 values in the mask are used as the coefficients here. We apply a softmax transform to make sure these 9 coefficients sum up to be one.

\section{Supplementary results}

\subsection{Full data tables}

We provide the full data table of all our experiments on the validation set in \cref{tab:tab123_app,tab:tab4_app,tab:tab5_app,tab:tab6_app}. For test sets, we have submitted test results to the benchmark website, so please refer to the website for full evaluations.

\setlength{\tabcolsep}{4.5pt}
\begin{table*}[tb]
\begin{center}
\begin{tabular}{cccc|ccccc|ccccc}
\hline
\multirow{2}{*}{up} &
  \multirow{2}{*}{no sm} &
  \multirow{2}{*}{enc} &
  \multirow{2}{*}{aug} &
  \multicolumn{5}{c|}{KITTI-2015} &
  \multicolumn{5}{c}{KITTI-2012} \\
 &
   &
   &
   &
  \underline{Fl\_all} &
  Fl\_noc &
  EPE\_all &
  EPE\_noc &
  EPE\_occ &
  \underline{Fl\_all} &
  Fl\_noc &
  EPE\_all &
  EPE\_noc &
  EPE\_occ \\ \hline
    &     &   &     & 10.360 & 8.528 & 2.901 & 2.068 & 6.967 & 5.707 & 3.741 & 1.406 & 0.886 & 4.417 \\ \hline
    &     & 1 &     & 9.920  & 8.087 & 2.685 & 1.885 & 6.568 & 5.484 & 3.606 & 1.354 & 0.861 & 4.205 \\
    &     & 2 &     & 9.830  & 7.951 & 2.646 & 1.857 & 6.396 & 5.450 & 3.592 & 1.339 & 0.860 & 4.115 \\
    &     & 3 &     & 9.745  & 7.977 & 2.608 & 1.853 & 6.543 & 5.359 & 3.511 & 1.328 & 0.850 & 4.098 \\
    &     & 4 &     & 9.745  & 7.951 & 2.644 & 1.852 & 6.553 & 5.437 & 3.564 & 1.343 & 0.852 & 4.183 \\ \hline
\cmark &     &   &     & 10.220 & 8.221 & 2.825 & 1.970 & 6.748 & 5.728 & 3.725 & 1.420 & 0.888 & 4.495 \\
\cmark & \cmark &   &     & 8.871  & 7.142 & 2.605 & 1.849 & 6.037 & 5.314 & 3.374 & 1.386 & 0.876 & 4.347 \\ \hline
\cmark & \cmark & 3 &     & 8.260  & 6.577 & 2.415 & 1.729 & 5.794 & 4.964 & 3.111 & 1.291 & 0.825 & \textbf{4.007} \\
\cmark & \cmark &   & \cmark & 8.801  & 6.748 & 2.484 & 1.595 & 6.642 & 5.421 & 3.281 & 1.411 & 0.852 & 4.653 \\
\cmark & \cmark & 3 & \cmark & \textbf{7.788}  & \textbf{5.963} & \textbf{2.179} & \textbf{1.399} & \textbf{5.635} & \textbf{4.872} & \textbf{2.932} & \textbf{1.284} & \textbf{0.788} & 4.175 \\ \hline
\end{tabular}
\end{center}
\caption{Full validation results for Table 1, 2, \& 3 in the main paper (EPE/px and Fl/\%). Metrics evaluated at `all' (all pixels, default for EPE), `noc' (non-occlusions), `occ' (occlusions), `bg' (background), and `fg' (foreground). Key metrics (used in official ranking) are underlined. `Ours (baselinse)'= up + no sm; `Ours (+enc)'= up + no sm + enc=3; `Ours (+enc +aug)' = `Ours (final)' = up + no sm + enc=3 + aug. For all metrics, lower is better.}
\label{tab:tab123_app}
\end{table*}
\setlength{\tabcolsep}{6pt}

\setlength{\tabcolsep}{4.5pt}
\begin{table*}[tb]
\begin{center}
\begin{tabular}{c|ccccc|ccccc}
\hline
\multirow{2}{*}{Options of aug} & \multicolumn{5}{c|}{KITTI-2015}                    & \multicolumn{5}{c}{KITTI-2012}                     \\
                                & \underline{Fl\_all} & Fl\_noc & EPE\_all & EPE\_noc & EPE\_occ & \underline{Fl\_all} & Fl\_noc & EPE\_all & EPE\_noc & EPE\_occ \\ \hline
Ours (final)                    & \textbf{7.788}   & \textbf{5.963}   & \textbf{2.179}    & 1.399    & 5.635    & \textbf{4.872}  & \textbf{2.932}   & \textbf{1.284}    & \textbf{0.788}    & \textbf{4.175}    \\ \hline
start from 100k             & 7.916   & 6.013   & 2.186    & \textbf{1.396}    & \textbf{5.406}    & 4.984   & 2.998   & 1.302    & 0.803    & 4.984    \\
vehicles only               & 7.940   & 6.110   & 2.212    & 1.435    & 5.781    & 4.950   & 3.041   & 1.304    & 0.804    & 5.571    \\
loss on new occ             & 8.149   & 6.105   & 2.272    & 1.418    & 5.620    & 5.167   & 3.083   & 1.361    & 0.810    & 5.892    \\ \hline
\end{tabular}
\end{center}
\caption{Full validation results for Table 4 in the main paper (EPE/px and Fl/\%).}
\label{tab:tab4_app}
\end{table*}
\setlength{\tabcolsep}{6pt}

\setlength{\tabcolsep}{4.5pt}
\begin{table*}[tb]
\begin{center}
\begin{tabular}{c|cccccccccc}
\hline
 &
  \textbf{road} &
  \textbf{car} &
  \textbf{terrain} &
  \textbf{vegetation} &
  \textbf{sidewalk} &
  \textbf{building} &
  \textbf{wall} &
  \textbf{pole} &
  \textbf{fence} &
   \\ \hline
Proportion              & 42.4\% & 17.6\% & 14.0\% & 11.6\% & 6.2\%  & 4.1\%  & 1.3\%   & 1.1\%   & 0.9\%  &   \\ \hline
ARFlow~\cite{liu2020learning}                  & 4.573   & 15.791  & 9.420   & 15.339  & 4.606   & 6.000   & 16.336   & 10.745   & 46.297  &    \\
Ours (final)            & 3.674   & 10.171  & 8.737   & 13.470  & 3.085   & 4.275   & 11.180   & 9.437    & 39.120  &   \\ \hline
Rel. impr.    & 19.7\% & 35.6\% & 7.3\%  & 12.2\% & 33.0\% & 28.8\% & 31.6\%  & 12.2\%  & 15.5\% & \\
Rew. contri. & \textcolor{red}{19.0\%} & \textcolor{red}{49.4\%} & 4.8\%  & \textcolor{red}{10.9\%} & 4.7\%  & 3.5\%  & 3.4\%   & 0.7\%   & 3.3\%  &   \\ \hline \hline
 &
 \textbf{traffic sign} &
  \textbf{truck} &
  \textbf{bicycle} &
  \textbf{traffic light} &
  \textbf{person} &
  \textbf{rider} &
  \textbf{motorcycle} &
  \textbf{sky} &
  \textbf{bus} &
  \textbf{train}
   \\ \hline
Proportion           & 0.4\%   & 0.2\%  & 0.1\%  & 0.1\%  & 0.1\%  & \textless0.1\%  & \textless0.1\%  & \textless0.1\%   & \textless0.1\%   & \textless0.1\%          \\ \hline
ARFlow~\cite{liu2020learning}         & 5.254         & 10.461  & 5.421   & 2.290   & 3.852   & 20.811  & 13.385  & 28.547   & 17.179   & 12.564           \\
Ours (final)        & 4.681     & 9.526   & 4.945   & 2.060   & 2.627   & 19.581  & 3.620   & 38.521   & 17.747   & 9.615            \\ \hline
Rel. impr.  & 10.9\%   & 8.9\%  & 8.8\%  & 10.0\% & 31.8\% & 5.9\%  & 73.0\% & -34.9\% & -3.3\%  & 23.5\%         \\
Rew. contri. & 0.1\% & 0.1\%  & \textless0.1\%  & \textless0.1\%  & \textless0.1\%  & \textless0.1\%  & \textless0.1\%  & \textless0.1\%  & \textless0.1\% & \textless0.1\%          \\ \hline
\end{tabular}
\end{center}
\caption{Full data for Table 5 in the main paper. We show the relative improvements and reweighted contributions of all 19 classes.}
\label{tab:tab5_app}
\end{table*}
\setlength{\tabcolsep}{6pt}

\setlength{\tabcolsep}{4.5pt}
\begin{table*}[tb]
\begin{center}
\begin{tabular}{c|ccccc|ccccc}
\hline
\multirow{2}{*}{Method} & \multicolumn{5}{c|}{KITTI-2015}                    & \multicolumn{5}{c}{KITTI-2012}                     \\
                        & \underline{Fl\_all} & Fl\_noc & EPE\_all & EPE\_noc & EPE\_occ & \underline{Fl\_all} & Fl\_noc & EPE\_all & EPE\_noc & EPE\_occ \\ \hline
ARFlow (our impl.)      & 13.210  & 10.190  & 4.081    & 2.878    & 9.398    & 7.360   & 4.434   & 1.713    & 0.994    & 5.806    \\
Ours (baseline)         & 12.270  & 8.642   & 3.809    & 2.433    & 9.907    & 7.165   & 4.104   & 1.730    & 0.997    & 5.915    \\
Ours (+enc)$^\dagger$             & 11.280  & 7.749   & 3.327    & 2.121    & 8.749    & 6.583   & 3.596   & 1.561    & 0.912    & 5.267    \\
Ours (+enc +aug)$^\dagger$        & \textbf{10.320}  & \textbf{6.950}   &\textbf{2.640}    & \textbf{1.558}    & \textbf{7.121}    & \textbf{6.204}   & \textbf{3.488}   & \textbf{1.489}    & \textbf{0.855}    & \textbf{5.150}    \\ \hline
\end{tabular}
\end{center}
\caption{Full validation results for Table 6 in the main paper (EPE/px and Fl/\%). $^\dagger$ denotes models with semantic inputs.}
\label{tab:tab6_app}
\end{table*}
\setlength{\tabcolsep}{6pt}

\subsection{More qualitative results}

\begin{figure*}
\begin{center}
    \includegraphics[width=\linewidth]{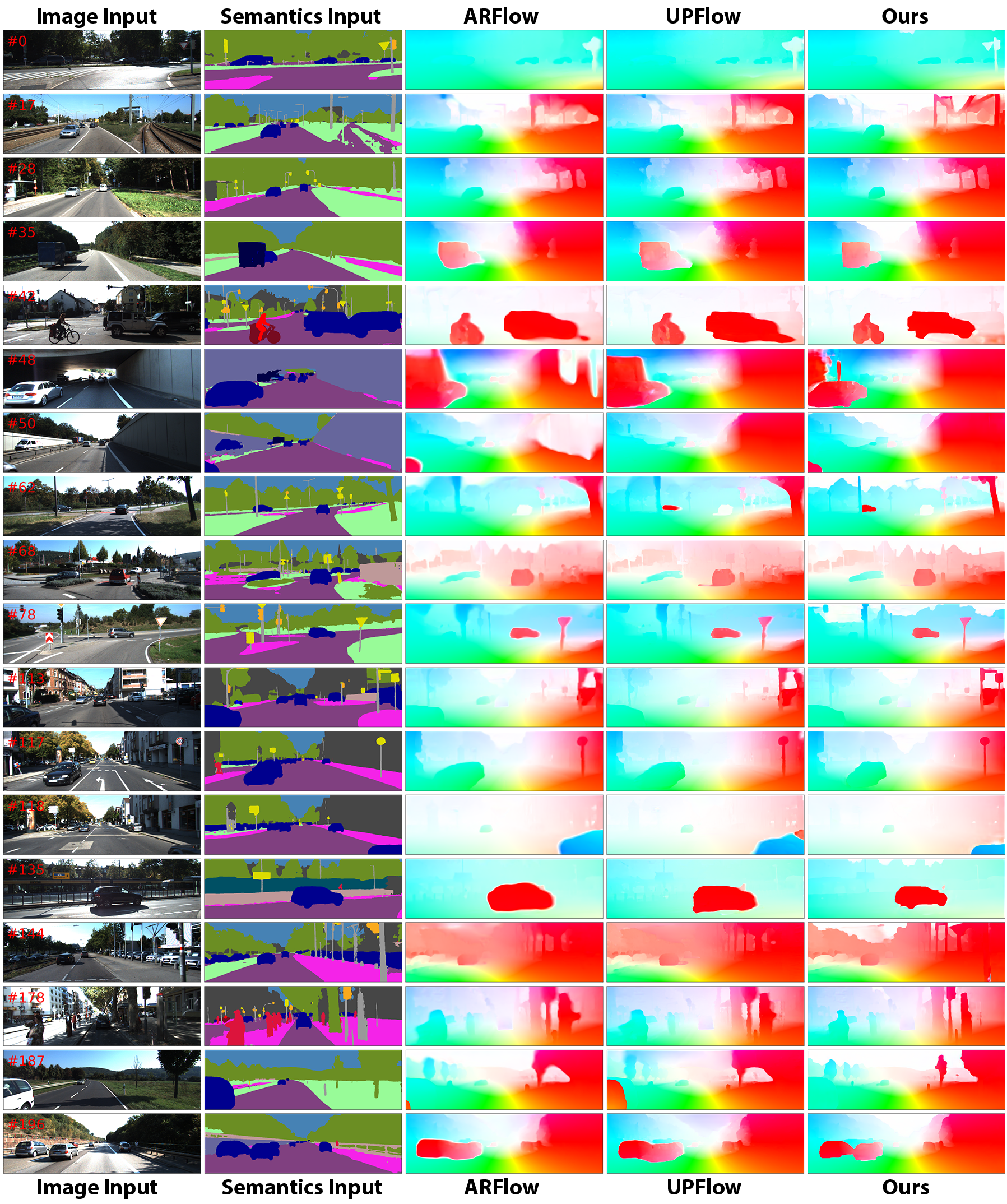}
    \caption{More qualitative results from the KITTI-2015 test set. Sample IDs are shown on the top left corners of the images.}
    \label{fig:qual_app}
\end{center} 
\end{figure*}

More qualitative results on the KITTI-2015 test set are shown in \cref{fig:qual_app}.

\subsection{Time efficiency}\label{subsec:time}

Our network runs very efficiently thanks to its small size. Our network with semantic encoder and learned upsampler only has 2.6M parameters in total, so the model parameter size is only around 10MB.

\paragraph{Training} We run 200k iterations in total. For our basic network with a semantic encoder and a learned upsampler, it takes around 44-48 hours on 2 NVIDIA GeForce RTX 2080 Ti GPUs. After adding semantic augmentation, it takes longer because we add a third forward pass of the network, but since we only use semantic augmentation for the last 50k iterations, the running time is still feasible: 54-58 hours on 2 NVIDIA GeForce RTX 2080 Ti GPUs.

\paragraph{Inference} For inputs of size $256\times 832$, inferring the forward flow of each sample takes 0.0168$ (\pm0.0005)$ second, \ie 60 frames per second, on a single NVIDIA GeForce RTX 2080 Ti GPU.

\paragraph{Tips on how to process semantic inputs efficiently} We need to one-hot encode the semantic map input before feeding into the network. We do one-hot transformation \emph{after} data augmentation to save time because otherwise, doing horizontal flip or bilinear interpolation for a 19-channel map is very time-consuming. We also avoid flipping or rescaling these 19-channel semantic maps when we copy and paste occluder objects across samples for semantic augmentation.

\clearpage
\newpage
\section{Other explorations}

We have also explored many other methods to apply semantics in the unsupervised optical flow network. Although these trials are not very successful and thus are not proposed in our final model, we briefly discuss our findings for the readers' reference.

\subsection{Adding semantic inputs to the decoder} 
We also tried adding semantic inputs to the decoder. This is because the current semantic maps are used as encoder inputs, which are somewhat distant from our final output, so we were wondering whether adding semantic cues to the decoder directly could help it decode a better flow. 

We used two shallow convolutional layers to extract a feature map from semantic input and downsample that feature to different resolutions for different levels of the iterative decoder. We found that such direct injection of semantic input into the decoder did improve the vanilla ARFlow without semantic encoder. However, it helped little for our adapted ARFlow model \emph{with} semantic encoder.

\subsection{Adding a semantic consistency loss}

As applied in some previous work~\cite{bai2016exploiting}, the semantic consistency loss enforces the output correspondence to have consistent semantic classes. This is similar to the photometric loss, but we use the output flow to warp the semantic inputs instead of the image inputs. 

After experimenting this semantic consistency loss, unfortunately, we did not find it very helpful to our network. There are mainly two reasons.

Firstly, unlike the image input, which consists of roughly continuous RGB values, the semantic input is categorical, and it is common to have large areas of the same semantic class. Thus, one may understand the semantic map as a highly texture-less channel of input, which cannot help differentiate different regions of the same semantic class. Also, since a large area has the same semantic class, tuning flow in that region makes no difference to the semantic consistency loss, which means the gradient will be zero for most pixels except for those near the semantic boundaries.

One solution to the aforementioned problem is to use a continuous semantic class distribution as input. For example, we can use the softmax values from the semantic segmentation network as our semantic input. However, as we mentioned earlier in \cref{subsec:time}, doing augmentations on a 19-channel semantic input is very time-consuming.

\begin{figure}[t]
\begin{center}
    \includegraphics[width=\linewidth]{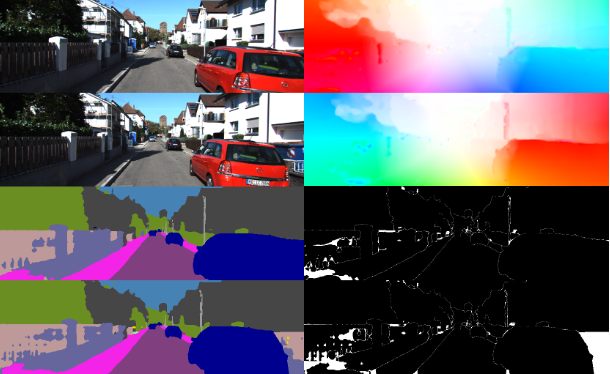}
    \caption{Left: two input frames and the semanic inputs; top right: forward and backward flow; bottom right: forward and backward semantic consistency loss}
    \label{fig:sem_cons_demo}
\end{center}
\end{figure}

Secondly, semantic consistency loss also does not work on occlusion regions, where photometric loss has issues, so we have to mask out the occlusion regions for both losses. For non-occluded regions, the current photometric loss is already good enough to find semantically consistent output by itself, so semantic consistency does not add much here. As illustrated in \cref{fig:sem_cons_demo}, we trained a model with only photometric loss for only 50k iterations, and most part of the frame is already very consistent on semantics. The inconsistent parts are mostly either on semantic boundaries or in the occlusion region.

\subsection{Using semantic boundaries for the boundary-aware smoothness loss}

Most previous methods use smoothness loss to constrain a smooth flow output. However, motion is not smooth across motion boundaries, where motion changes abruptly. Motion boundaries usually coincide with object boundaries, so object boundaries can be a good approximation to indicate where smoothness loss should not be imposed.

Due to lack of semantic information, current methods use image edges instead to generate a weight map to reweigh smoothness loss at different pixels. In our case, since semantic maps are available, we use this information to create much clearer object boundaries. 

\begin{figure}[!ht]
\begin{center}
    \includegraphics[width=\linewidth]{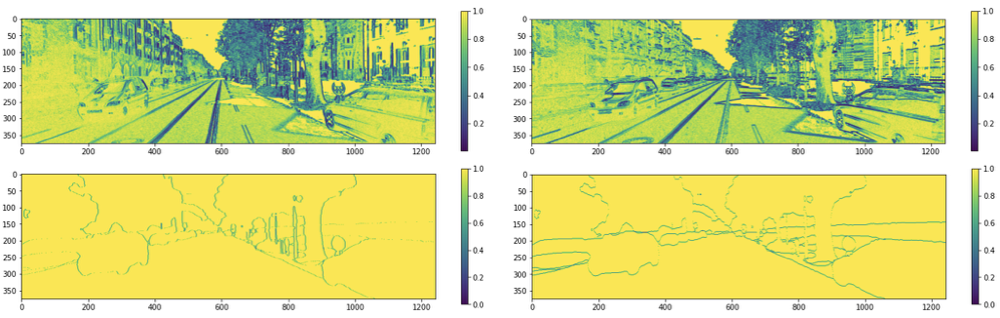}
    \caption{Top: weights based on image edges; bottom: weights based on semantic boundaries}
    \label{fig:sm_edge}
\end{center}
\end{figure}

We start from the same smoothness loss in ARFlow~\cite{liu2020learning}. As visualized in \cref{fig:sm_edge}, the weights based on image edges are computed as the sum of the second-order image derivatives on both x and y-axis, \ie, the Laplacian of the 2D optical flow field. In comparison, our semantic boundaries are much cleaner.

However, both image edges and semantic boundaries have issues. Image edges usually provide boundaries within the same object, such as the edge of the shadow on the road, and they also have fewer boundaries in the dark region where image values are similar. Meanwhile, semantic boundaries are computed from semantic segmentation, where different instances of the same semantic class are not differentiated. This causes big issues when, for example, the semantic map of multiple cars merge into one.

\begin{figure}[!ht]
\begin{center}
    \includegraphics[width=\linewidth]{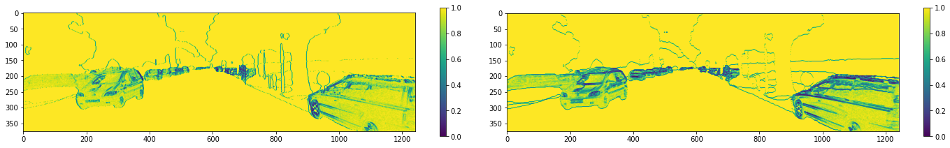}
    \caption{The combined boundary weight}
    \label{fig:sm_comb_edge}
\end{center}
\end{figure}

To fix these issues, we attempted to use a combination of both image edges and semantic boundaries. We find image edges in the vehicle (car, truck, bus, train), people (person, rider), and small vehicle (motorcycle, bicycle) regions, combined with semantic boundaries else where, as shown in \cref{fig:sm_comb_edge}.

We tried all these boundaries and tuned the weight of the smoothness loss term in the total loss, and found little difference in the evaluated results. Moreover, after we add the learned upampler module in the network, applying smoothness loss is doing more harm than help. This is understandable because the goal of the learned upsampler is to make the network decide which part of the flow output should be smooth and where should not be smooth. Smoothness loss imposes a preference that the motion should be smooth in the form of zero second-order derivatives, which is not data-driven and may not be precise for real-world motion fields. Therefore, we ended up turning off the smoothness loss in our final model.

\subsection{Learning the initial flow in the decoder}

Driving scenes usually have similar scene layouts (sky is on the top, and road is on the bottom, \etc). In addition, the motion of the camera is also mostly moving forward with some occasional slight rotations. These two effects together create a \emph{looming} motion, where most objects are moving closer to the camera. To explain in 2D image frames, the left part of the image tends to move left, and the right part tends to move right. The lower part (mostly roads and sidewalks) also tends to move downwards until they are out of sight. These observations indicate a strong motion prior knowledge that can be used to better initialize our flow estimate.

The current iterative decoder as in ARFlow~\cite{liu2020learning} initialize the flow estimate as zero motion, which gets refined later iteration by iteration. However, we can apply the looming motion prior instead by parameterizing the initial flow using some learnable parameters. Specifically, for a $256\times832$ input, the highest (6th) level feature has dimension $4\times 13$, so we use a learnable $19\times 4\times 13 \times 2$ tensor as the prior. We condition the motion prior on the 19 semantic classes, so that we can refer to the semantic map input to generate its initial flow prior based on semantics. Note that we define the prior specifically forward flow only, and we need to flip the prior when computing backward flow in our network.

\begin{figure}[!ht]
\begin{center}
    \includegraphics[width=\linewidth]{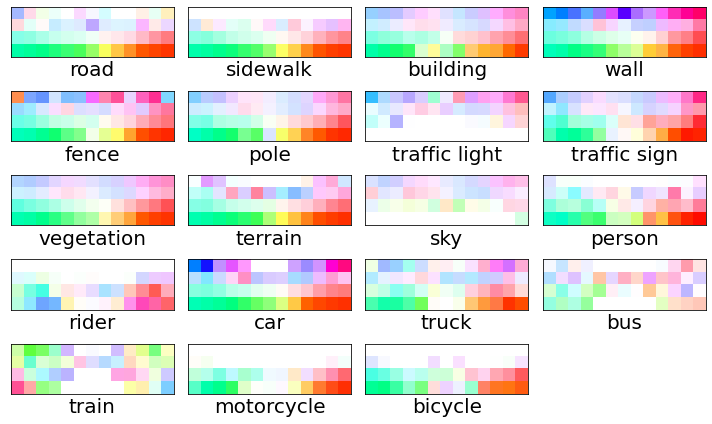}
    \caption{The learned init flow prior for each semantic class}
    \label{fig:init_flow_prior}
\end{center}
\end{figure}

We trained the network with learned initial flow conditioned on semantics, and the learned prior is visualized in \cref{fig:init_flow_prior}. Overall, we can see that the looming motion pattern is learned by our network. However, for each semantic class, the network can only learn prior for places where that class frequently appears. For instance, the upper part of the road prior is not learned well because there are few road pixels on the top of the frame.

\begin{figure}[!ht]
\begin{center}
    \includegraphics[width=\linewidth]{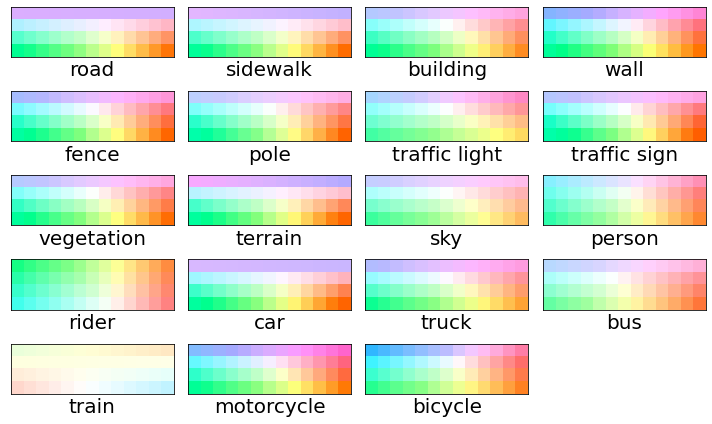}
    \caption{The learned init flow prior for each semantic class if we parameterize by coners}
    \label{fig:init_flow_prior_corner}
\end{center}
\end{figure}

To better fix this issue, we try to parameterize the motion prior by its four corners. Specifically, we learn a $2\times2$ prior for each class, and bilinear interpolate this $2\times 2$ prior to $4\times 13$ before using it to construct initial flow based on semantics. The results are then visualized in \cref{fig:init_flow_prior_corner}. The network has learned a very smooth and more or less similar pattern prior for each class.

In terms of evaluation metrics, both methods improve slightly by themselves, but we found the improvements became negligible after we apply the semantic segmentation module. One question here is whether good initialization matters a lot for our unsupervised flow networks. Since we refine the flow by many iterations in the decoder, the initial estimate may not change the results significantly, if the following refinement units are effective enough.

\subsection{Reweighing losses at different semantic regions}

Current models usually have different scales of error at different semantic regions. Semantic classes such as car are harder to track and thus incur larger errors than other classes. Also, some classes, such as car and person, are practically more important for autonomous driving applications. Therefore, we tried to reweigh the photometric loss by their semantic class. We give higher weights to classes like car and person so that the network can focus on improving those classes more.

However, the results are hard to evaluate numerically. The current ground-truth labels are mostly concentrated on the lower part of the frame, so many practically important objects such as traffic lights, traffic signs and poles only account for a very small amount of the evaluation. Also, classes like person, rider, and bicycles do not have flow labels because reliable CAD models are not available for these dynamic objects. Based on these issues, finding a better evaluation method may be more important.

\subsection{Using the epipolar constraint to post-process the static region flow}

Following earlier traditional methods~\cite{hur2016joint,wulff2017optical}, we also explored the possibility of using the static scene epipolar constraint to post-process static optical flow. Since most part of the frame is static, we can use our correspondences found to estimate the fundamental matrix between two frames, and then use epipolar constraints to refine our flow.

This method is mostly targeted on refining optical flow for those occluded static pixels. For example, a part of the road or background may be occluded by the moving cars, or they may simply move out of the frame, so their correspondence is not visible in the other frame. However, we still want the network to have a best ``guess'' on where the correspondence is. Most current methods rely on smoothness to generate those ``guesses''. However, given the epipolar constraint, we limit the searching range of correspondence to one epipolar line, which may help us ``guess'' more informatively. Our semantic map inputs tell us where those static region is, so we can get a more reliable fundamental matrix estimate. 

Estimating the fundamental matrix only requires eight pairs of corresponding points, so we can select only the most reliable correspondences for this computation. Specifically, we define ``reliable correspondence'' based on three criteria: (1) not in the occlusion region, (2) not on the semantic boundary or image borders, and (3) not from any (possibly) dynamic semantic class or any texture-less semantic class.

\begin{figure}[!ht]
\begin{center}
\subfigure[Inputs for post-processing]{
    \includegraphics[width=\linewidth]{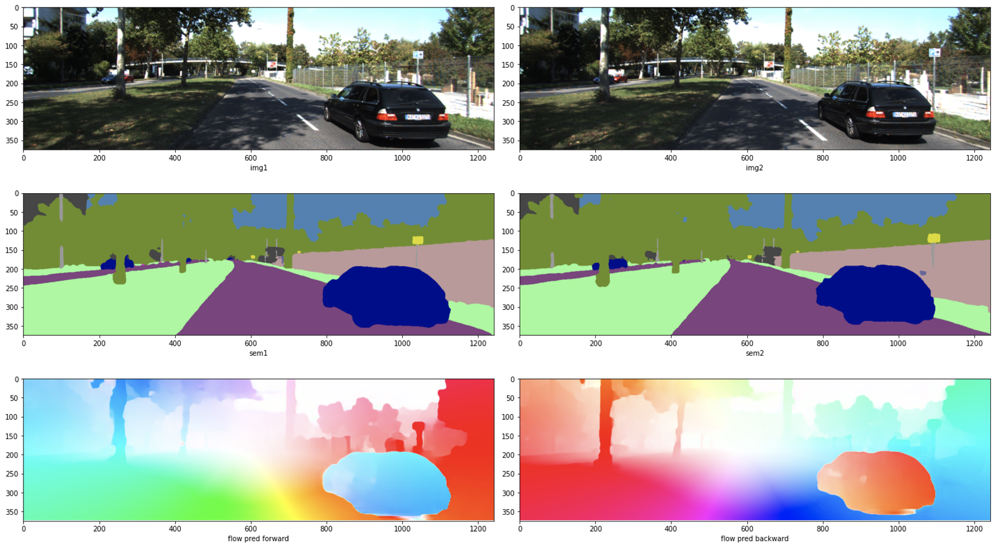}
    \label{fig:post1}
    }
\subfigure[Computing reliable static region masks]{
    \includegraphics[width=\linewidth]{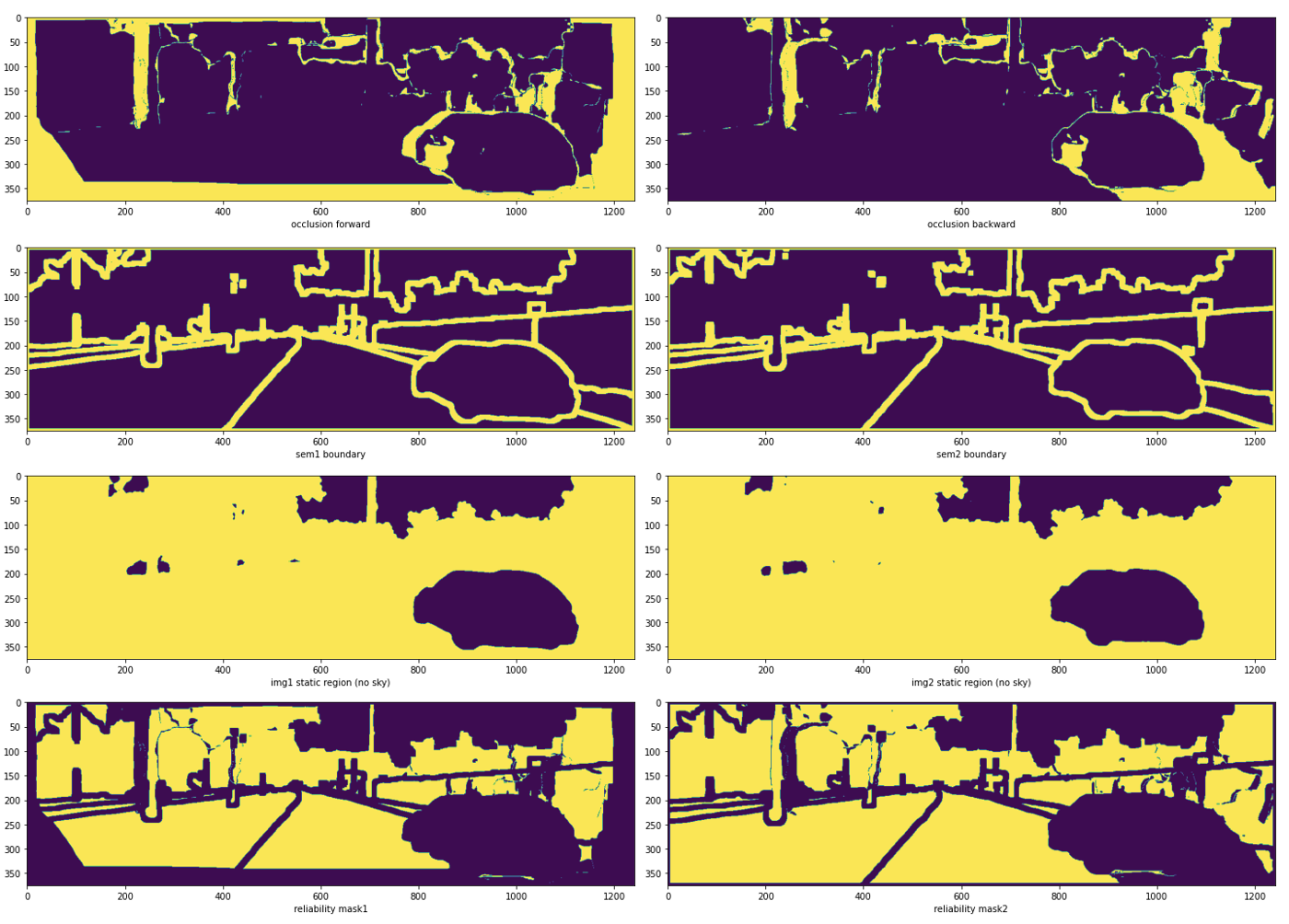}
    \label{fig:post2}
    }
\subfigure[Comparing epipolar errors and true errors]{
    \includegraphics[width=\linewidth]{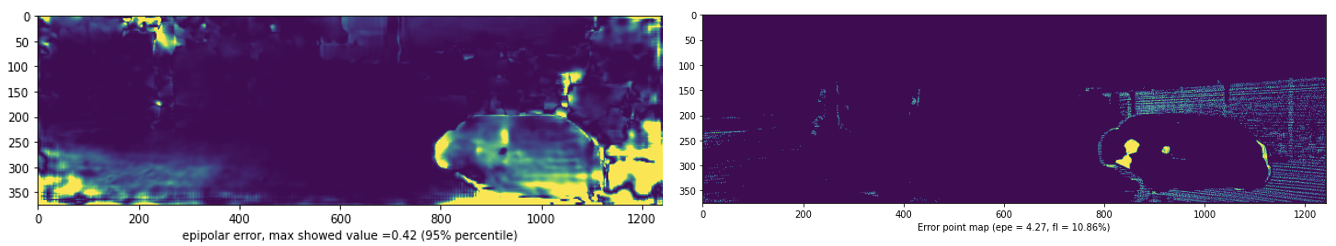}
    \label{fig:post3}
    }
\end{center}
\caption{An example for semantic post-processing}
\label{fig:post}
\end{figure}

We first compute the occlusion mask through forward-backward consistency check. Then, we find semantic boundaries based on our semantic map input and define all pixels that are  $\leq 5$ pixels away from any boundary as the boundary pixels. For reliable static semantic classes, we include all classes except vehicles (car, truck, bus, train), people (person, rider), small vehicles (bicycle, motorcycle), and sky (poor texture). One example is shown in \cref{fig:post1,fig:post2}.

After computing the reliable static regions, we use both forward and backward flow in those regions to create a set of correspondences between two input frames. We then estimate the fundamental matrix $\hat F$ using RANSAC. For a typical sample in the KITTI-2015 train set, RANSAC based on correspondeces in our reliable regions mostly produces \textgreater98\% inlier rate.

We then find the correspondences in static regions and check whether they conform to epipolar constraint. For each point $\bm p$ in the static region of the first frame, we compute the epipolar error
\[
\epsilon(\bm p) = \left((\bm p + U_{1\to2}(\bm p))^T F \bm p\right)^2,
\]
where $U_{1\to2}(\bm p)$ is the input flow estimate. We find the pixels with top 5\% epipolar error and refine those by projecting their current estimated correspondences onto their epipolar lines. As shown in \cref{fig:post3}, our epipolar error successfully detects the part of the frame that has high EPE errors. However, we tested this method on the KITTI-2015 train set but did not see much improvements on the evaluation results. The reason is still under investigation.

\end{document}